\documentclass[journal]{IEEEtran}

\usepackage{multicol}
\usepackage{color}
\usepackage{xspace}
\usepackage{times}
\usepackage{epsfig}
\usepackage{graphicx}
\usepackage{amssymb}
\usepackage{amsmath}
\usepackage{url}
\usepackage{multirow}
\newcommand{\ryn}[1]{\textcolor{black}{#1}}

\newcommand{\fir}[1]{\textcolor{red}{#1}}
\newcommand{\se}[1]{\textcolor{blue}{#1}}
\newcommand{\thi}[1]{\textcolor{green}{#1}}

\begin{document}
\title{Deformable Object Tracking with Gated Fusion}

\author{Wenxi~Liu, \ \
		Yibing~Song, \ \
		Dengsheng~Chen, \ \
		Shengfeng~He, \\
		Yuanlong~Yu, \ \
		Tao~Yan, \ \
        Gerhard P. Hancke, \ \ and \ \
		Rynson~W.H. Lau
\IEEEcompsocitemizethanks{\IEEEcompsocthanksitem Wenxi Liu, Dengsheng Chen, and Yuanlong Yu are with the College of Mathematics and Computer Science, Fuzhou University, China. \protect
\IEEEcompsocthanksitem Yibing Song is with the Tencent AI Lab, Shenzhen, China.\protect
\IEEEcompsocthanksitem Tao Yan is with the Jiangsu Key Laboratory of Media Design and Software Technology, Jiangnan University, China.\protect
\IEEEcompsocthanksitem Shengfeng He is with the School of Computer Science and Engineering, South China University of Technology, China.\protect
\IEEEcompsocthanksitem Gerhard P. Hancke and Rynson W.H. Lau are with City University of Hong Kong, Hong Kong.\protect
\IEEEcompsocthanksitem Shengfeng He and Yuanlong Yu are the corresponding authors. This project is led by Rynson W.H. Lau. 
} 
\thanks{}}

	\markboth{Journal of \LaTeX\ Class Files,~Vol.~14, No.~8, August~2015}%
	{Shell \MakeLowercase{\textit{et al.}}: Bare Demo of IEEEtran.cls for IEEE Journals}
	%

	\maketitle
	
	\begin{abstract}
		The tracking-by-detection framework receives growing attentions through the integration with the Convolutional Neural Networks (CNNs). Existing tracking-by-detection based methods, however, fail to track objects with severe appearance variations. This is because the traditional convolutional operation is performed on fixed grids, and thus may not be able to find the correct response while the object is changing pose or under varying environmental conditions. In this paper, we propose a deformable convolution layer to enrich the target appearance representations in the tracking-by-detection framework. We aim to capture the target appearance variations via deformable convolution, which adaptively enhances its original features. In addition, we also propose a gated fusion scheme to control how the variations captured by the deformable convolution affect the original appearance. The enriched feature representation through deformable convolution facilitates the discrimination of the CNN classifier on the target object and background. Extensive experiments on the standard benchmarks show that the proposed tracker performs favorably against state-of-the-art methods.
	\end{abstract}
	

	%
	\IEEEpeerreviewmaketitle

	\section{Introduction}
Visual object tracking is one of the fundamental problems in computer vision and has many applications, e.g., security surveillance, autonomous driving, and human-computer interactions. In recent years, with the advancement of deep convolutional neural networks (CNNs), which can extract features that are more discriminative than the empirical ones, visual tracking has achieved favorable performance on multiple standard benchmarks.
	
\begin{figure}
		\centering
		\includegraphics[width=0.5\textwidth]{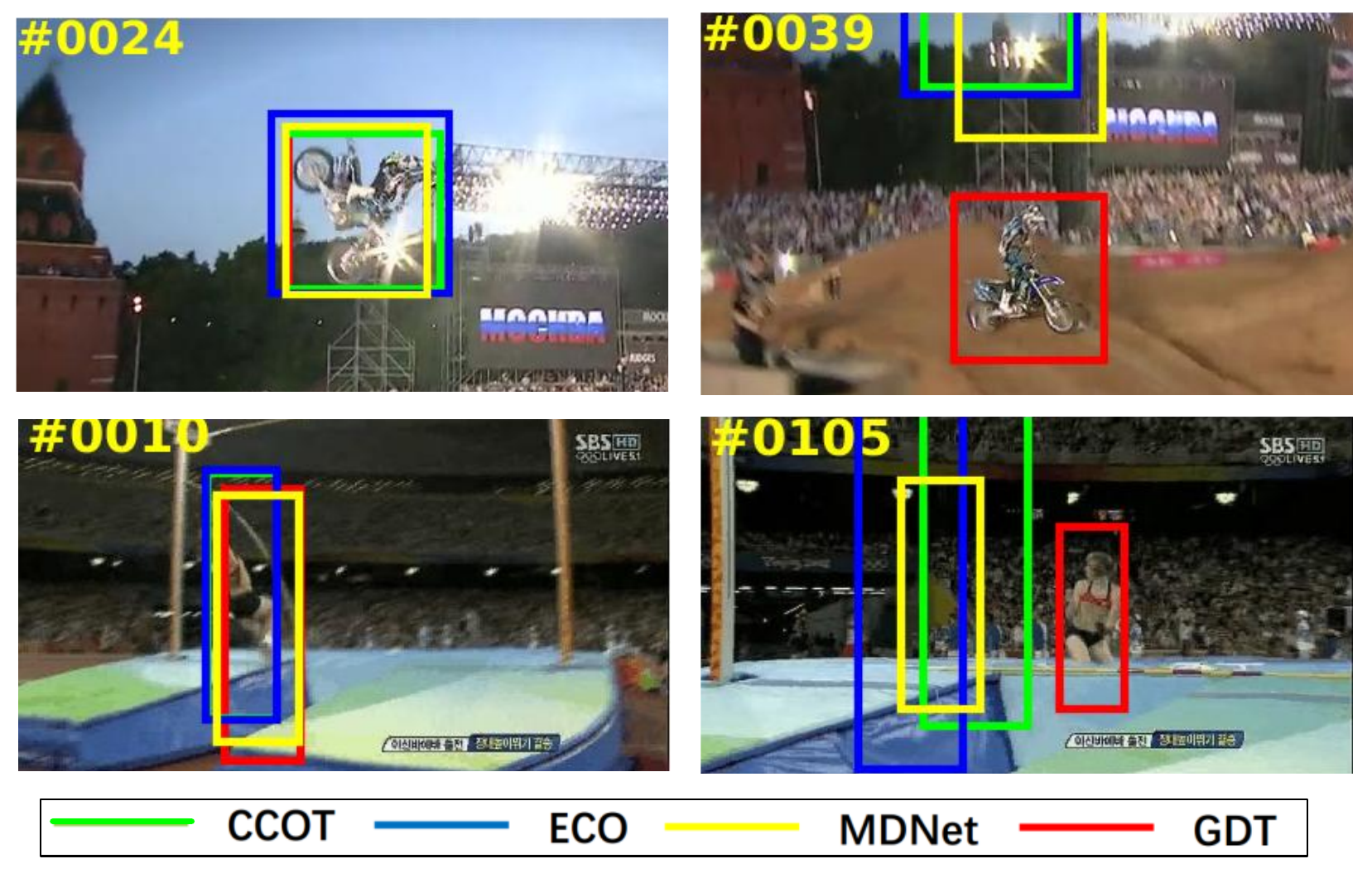}
		
		\caption{Illustration of significant appearance variations (i.e., rotation and deformation) in videos \textit{MotorRolling} and \textit{Jumping}. Compared with the state-of-the-art methods, our gated-fusion deformable tracker (denoted as \textit{GDT}) can extract features from rapid appearance variations and correctly localize the targets.}
		
		\label{fig:teaser}
		
	\end{figure}

Despite the demonstrated success, existing state-of-the-art tracking methods suffer from large object appearance variations. The setting of visual tracking uses only the first frame as input, which contains limited representations of the target appearance. Tracking accuracy deteriorates when the target object undergoes severe appearance variations (e.g., pose variation, deformation, and rotation as shown in Fig.~\ref{fig:teaser}). Existing methods are designed without sufficient modeling of such severe appearance variations, degrading the classifier's ability to  discriminate the target object from the background.
	
As we have observed, the deformable objects do not always reside in the regular grids of the image space and the relative locations of the object parts often vary in video frames. However, the existing CNN-based tracking methods lack internal mechanisms to handle deformations, since the standard CNNs perform the convolution operation over a fixed geometric structure. In prior vision tasks, a common solution to this problem is to collect an extensive amount of training samples. Training data, however, is difficult to collect during online tracking. Hence, given only an input sample in the first frame, the normal convolutional features often fail to model the object with significant pose variation, deformation, or occlusion in tracking.

To tackle this problem, we present a deformable convolutional layer in the CNN-based tracking-by-detection framework in order to model the appearance variations.
The deformable convolutional layer enables free-form deformation of the sampling grid.  Thus, it can extract features adaptively according to the changing object appearances. In particular, when the target object undergoes severe appearance variations, the deformable convolution layer aims to generate a normal response similar to those in the ordinary scenarios by estimating the free-form deformation. In addition, although the training samples collected online are limited and similar, the deformable convolution is capable of adapting to unseen deformations via online learning.
	
	On the other hand, relying solely on the deformable convolution may have some limitations, e.g., the degradation of scale estimation and localization. This is because it may treat scaling and shifting as some forms of deformations, and thus try to recover those negative samples and categorize them as positive samples. We note that when the target object is in an ordinary scenario with minor appearance variations, the well-trained normal convolution features are effective. When the target object has significant appearance variations, the deformable convolution will be more effective. Hence, we introduce a soft gate mechanism to balance between the normal convolutional features and the deformable convolutional features.
	The soft gate approves the fusion between the deformable features and the original ones when the appearance variations of the target object are accurately modeled. Specifically, the soft gate adaptively blends these two types of features.
	As a result, the gated fusion of the normal convolutional features and the deformable ones will recover the rapid appearance variations of the target object into ordinary conditions. The gated fusion produces accurate deformable feature maps for the target object, which in turn enrich target appearances and facilitate the classifier prediction. In addition, the deformable convolutional layer with gated fusion is integrated into the tracking-by-detection framework for end-to-end training and prediction. Extensive experiments on the standard benchmarks indicate that the proposed tracker performs favorably against the state-of-the-art methods.

	We summarize the contributions of this work as follows:
	\begin{itemize}
		\item We present a deformable convolutional layer for the CNN-based tracking-by-detection framework to model target appearance variations.
		\item We propose a gated fusion mechanism to control the effect of the deformable convolutional layer on the output feature maps. It facilities the classifier's discrimination on the target object and background.
		\item Our tracker outperforms state-of-the-art approaches on the standard tracking benchmarks, especially on those challenging scenarios.
	\end{itemize}
	
	For the rest of the paper, we first survey related literatures in Section~\ref{sec:related_works}. We then present our deformable object tracker with gated fusion in Section~\ref{sec:method}. Finally, we evaluate the performance of our proposed tracker in several public benchmarks in Section~\ref{sec:exp}.

	\section{Related Works}
	\label{sec:related_works}

	In visual tracking, state-of-the-art trackers can be roughly categorized as: tracking-by-detection based methods, Siamese-network based methods, reinforcement-learning based methods, and deformable object trackers.

	\subsection{Tracking-by-detection Based Approaches}
	The tracking-by-detection framework considers the tracking task as a target/background classification problem.
	Numerous learning schemes have been proposed including P-N
	learning~\cite{kalal2012tracking}, online boosting~\cite{grabner2006real}, multiple instance learning~\cite{babenko2009visual}, structured SVMs~\cite{hare2016struck}, CNN-SVMs~\cite{hong2015online}, random forests~\cite{zhang2017robust}, domain adaptation~\cite{nam2016learning}, LSTM-based~\cite{Ning2017Spatially}, adversarial learning~\cite{song18cvpr}, reciprocative learning~\cite{shi2018DAT} and ensemble learning~\cite{han2017branchout}.
	Our proposed method is based on a CNN-based tracking-by-detection framework.
	Here, we focus on handling the challenging tracking task: tracking deformable objects while maintaining high-quality tracking performance in ordinary situations. To accomplish this,
	we introduce a gated fusion module to adaptively approve the fusion of the normal convolutional features and the deformable convolutional features.
	
{In recent years, the discriminative correlation filter (DCF) based trackers~\cite{bolme2010visual,henriques2015high} are widely studied. They gain much attention due to their real-time performance. In essence, they are related to tracking-by-detection methods, since they efficiently learn a discriminative regressor from foreground and background samples.
In particular, they regress all the circular-shifted samples into soft labels and transfer the correlation as an element-wise product in the Fourier domain. However, most of the DCF based methods suffer from the boundary effect and the model overfitting issues, which are caused by the circular-shifted samples and the dense training samples collection.
    To handle these issues, there are many extensions, including: kernelized correlation filters~\cite{henriques2015high}, scale estimation~\cite{danelljan2014accurate}, re-detection~\cite{ma2015long}, spatial regularization~\cite{danelljan2017eco,danelljan2016beyond,danelljan2015learning}, ADMM optimization~\cite{galoogahi2015correlation}, deep feature integrations~\cite{ma2015hierarchical,qi2016hedged,zhang2017multi,kiani2017learning,sun18cvpr}, and end-to-end CNN predictions~\cite{wang2015visual,song2017crest,valmadre2017end}.}s

\subsection{Siamese-Network Based Approaches}

Siamese-network based approaches have become popular in the visual tracking community, due to their balanced accuracy and speed~\cite{bertinetto2016fully,GOTURN_held16eccv,CFNet_Valmadre17cvpr,DSiam_qing17iccv,SiamRPN_bo18cvpr,RASNet_wang18cvpr}. {Unlike the tracking-by-detection methods, these methods formulate the object tracking as a similarity learning problem.} By comparing the ground-truth patch of the target object with the candidate patches within the search window at the current frame, the most similar patch is considered as the target. The similarity learning is accomplished by a fully convolutional Siamese network framework, which receives a pair of inputs and outputs the similarity. Most of these methods require little or no online training. Hence, they can reach real-time performance and are less affected by drifting caused by the online updating. However, they require a large amount of training data in the offline stage to achieve the state-of-the-art performance.

\subsection{Reinforcement-Learning Based Approaches}

Recently, researchers introduce deep reinforcement learning (DRL) into visual tracking. Deep reinforcement learning utilizes the deep neural network to model the active-value function to play games (e.g., Atari~\cite{Mnih2015Human}), which reaches human-level performance. As a potential research direction in tracking, prior works begin to use DRL to learn actions for robust tracking~\cite{ADNet_yun17cvpr,dong18cvpr,EAST_huang17iccv}, e.g., shifting and scaling the tracking window. The tracking related policy is learned from the training data in an offline manner.

	\begin{figure*}
		\centering
		
		\includegraphics[width=1.0\textwidth]{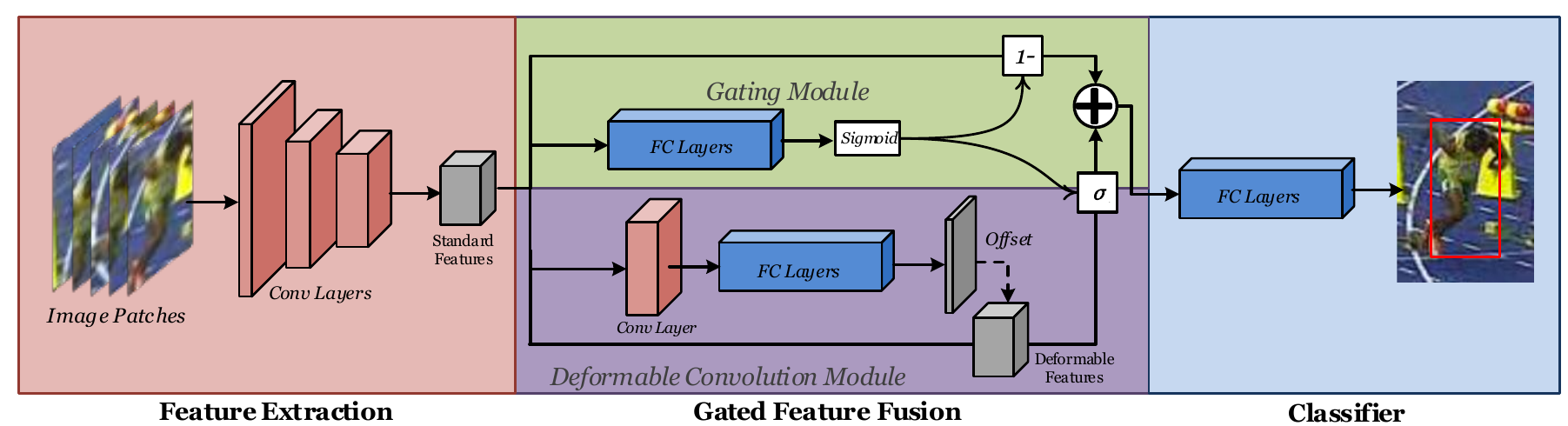}
		
		\caption{Our proposed framework is composed of three stages: (1) feature extraction consisting of pretrained convolutional layers, (2) gated feature fusion, and (3) classifier consisting of fully connected layers. Our proposed method focuses on the gated feature fusion, which includes a deformable convolution module, and a gating module that controls the fusion of the deformable features and the standard features.}
		

		\label{fig:framework}
		
	\end{figure*}

\subsection{Deformable Object Trackers}

{Tracking deformable objects is an important problem in visual tracking~\cite{zhao2016learning,du2016online,sun2017non,lukevzivc2017deformable,zhong2000object,leymarie1993tracking,gao2018p2t}. There are prior works focusing on tracking non-rigid deformable objects while segmenting their contours, including dynamic graph tracker \cite{Zhaowei2014Robust}, temporally coherent part-based tracker~\cite{Li2015Online}, local and global tracker~\cite{Luka2013Robust}, superpixel tracker~\cite{Shu2011Superpixel}, adaptive structural local sparse appearance model~\cite{Xu2012Visual}, latent structural learning tracker \cite{Yao2013Part}, sparsity-based collaborative model~\cite{Wei2012Robust}.
The general approach is to divide the object into parts. Local connectivity is typically applied among different parts of the object and a trade-off of the visual and geometric agreement is then optimized.
In \cite{sun2017non}, a shape-preserved kernelized correlation filter based tracker is proposed to use a set of deformable patches dynamically to collaborate on tracking of non-rigid objects.
In order to track deformable objects, Du et al. \cite{du2016online} capture the high-order temporal correlations among target parts by a structure-aware hyper-graph. }
However, the overall tracking performance of these methods is not comparable to the holistic methods. In this paper, we introduce deformable convolution that can both model the appearance of the deformable object as well as be incorporated into the CNN-based framework and trained in an end-to-end manner. Therefore, it can achieve state-of-the-art performance on public benchmarks.

\section{Deformable Tracker with Gated Fusion}
\label{sec:method}
Our approach is designed based on the tracking-by-detection framework, which includes the deformable convolution module and the gating module, as shown in Fig.~\ref{fig:framework}). {In particular, a set of candidate patches are randomly sampled from the current video frame and are fed to the network as input and its feature maps are obtained via several pretrained convolution layers (i.e., Conv layers) in the first stage. These standard feature maps are then sent to separate branches of the second stage. From Fig.~\ref{fig:framework}, for the bottom two branches, the deformable convolution module resamples the standard feature maps to produce the deformable convolutional features. On the other hand, the top branch of the gated feature fusion stage keeps the standard convolutional features, while the second branch is for the gating module to infer the weights that balance the effect of the standard convolutional features and the deformable features.} In the final stage, the fused features are sent to the fully connected layers (i.e., FC layers), which serve as a foreground/background classifier to detect the object in an online manner. In the following subsections, we introduce these modules and our tracking framework.

\subsection{Deformable Convolution}
\label{sec:feat}

\begin{figure}
	\centering
	
	\includegraphics[width=0.45\textwidth]{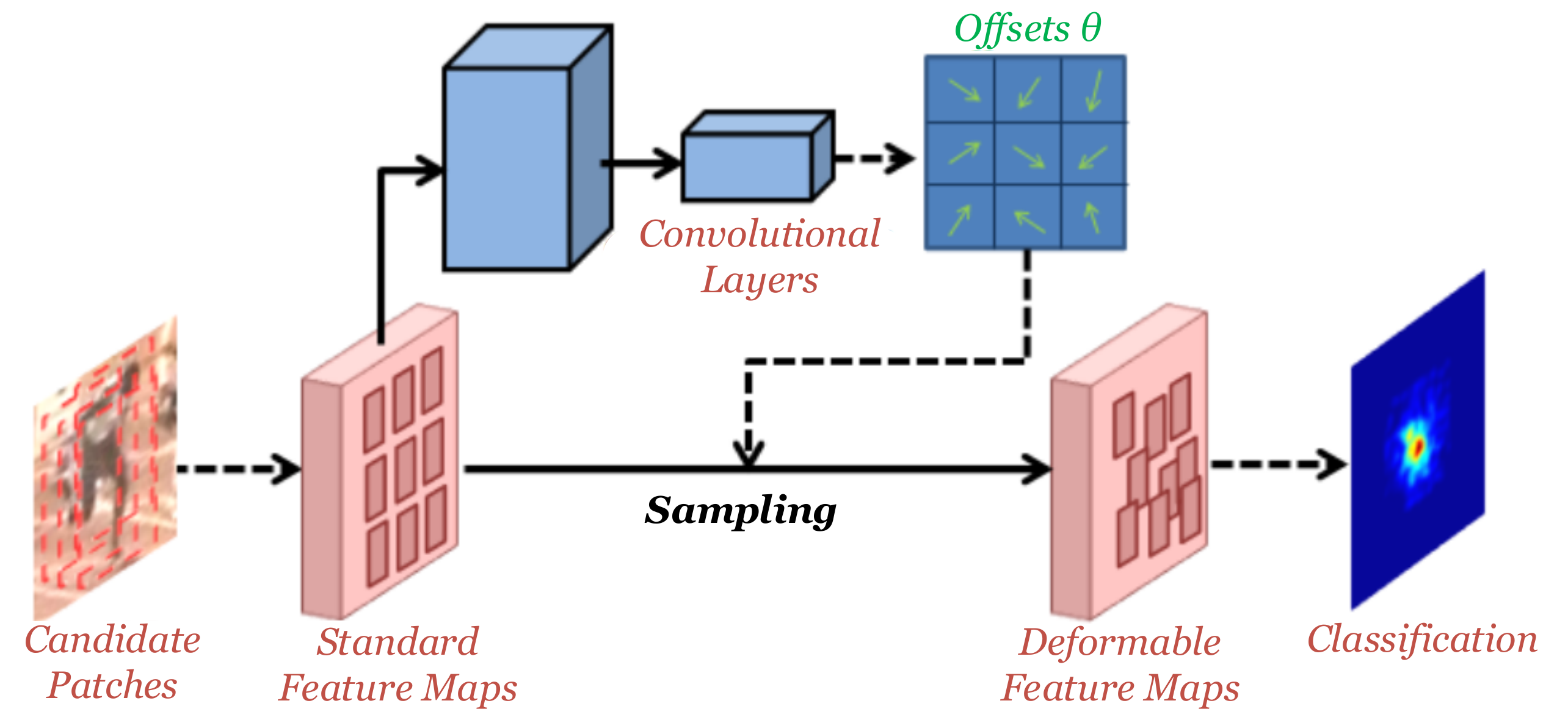}
	
	\caption{Illustration of the \textit{Deformable Convolution Module}. We demonstrate how the deformable features are computed. \ryn{The standard feature maps} are first passed through the convolutional layers to regress the deformation offsets $\Theta$. The offsets tell how the feature maps should be deformed and thus how they should be sampled.
	}
	
	\label{fig:deformnet}
	
\end{figure}

In order to model object variations in tracking, we introduce a deformable convolution module. As shown in Fig.~\ref{fig:framework}, the deformable convolution module is inserted in one of the branches of the CNN-based framework for extracting the deformable features. Fig.~\ref{fig:deformnet} illustrates how the deformable convolution module works.

To learn how to deform the feature maps, we are inspired by ~\cite{dai2017deformable}. As shown in Fig.~\ref{fig:deformnet}, the feature maps (the pink rectangle) are passed through an extra branch containing multiple convolutional layers (the blue rectangles) to regress the 2D offsets of the regular grid sampling locations. Then, the offsets are applied back to the feature maps and produce the new feature maps by resampling. {Formally}, the feature maps $\textbf{X}$ ($\textbf{X}\in \mathbb{R}^{H \times W \times C}$) are passed to the convolutional layers, which can be denoted as a non-linear function $\mathcal{F}_{deform}$. Thus, its output is reshaped and regresses the deformation offsets of the sampling locations $\Theta$ ($\Theta \in \mathbb{R}^{H \times W \times 2}$), as follows:
\begin{align}
\Theta = \mathcal{F}_{deform}(\textbf{X}),
\end{align}
where the convolutional layers consist of a convolutional layer with $3\times 3$ kernel size followed by a fully connected layer whose output has the same size as $\Theta$, i.e., $H*W*2$. In particular, $\theta_{i,j}$ ($\theta_{i,j} \in \Theta$) refers to the 2D offset vector that directs how the element $\textbf{X}_{i,j}$ ($\textbf{X}_{i,j} \in \textbf{X}$) in the standard feature maps deforms. Hence, the deformable feature maps are calculated as follows:
\begin{align}
\textbf{X}^\prime_{i,j} &= \mathcal{F}_{sample}(\textbf{X}, {[i,j]+ \theta_{i,j}}), \nonumber\\
\text{s.t.  } 	& \theta_{i,j} \in \mathbb{R}^{1 \times 1 \times 2}, \theta_{i,j} \in \Theta, \nonumber\\
& 1 \leq i \leq H, 1 \leq j \leq W,
\end{align}
\noindent where {$\textbf{X}^\prime_{i,j}$ are the deformable features at location $[i,j]$, and are sampled from location $[i,j] + \theta_{i,j}$, instead of location $[i,j]$}, with the shape of the offset vector $\theta_{i,j}$ transformed to $\mathbb{R}^{1\times 2}$ in advance. Note that $\theta_{i,j}$ is often fractional. So, a sampling kernel is applied to the feature maps as shown in Fig.~\ref{fig:deformnet}. Here, $\mathcal{F}_{sample}(\textbf{X},[d_x, d_y])$ is a bilinear interpolation kernel which samples $\textbf{X}$ at location $[d_x, d_y]$, as follows:
\begin{align}
\mathcal{F}_{sample}(\textbf{X},d)&=\sum_{x=1}^{W}\sum_{y=1}^H G([x,y],[d_x,d_y])\textbf{X}_{x,y}, \\
G([x,y],[d_x,d_y]) &= \max(0,1-|x-d_x|) \max(0,1-|y-d_y|),
\end{align}
\noindent where $\mathcal{F}_{sample}(\textbf{X},[d_x, d_y])$ samples the feature maps by calculating the weighted sum of the features at neighboring locations. $G(\textbf{a},\textbf{b})$ computes the weights of features at location $\textbf{a}$ for sampling the features at location $\textbf{b}$.


As we know, the standard CNNs perform the convolution operation over a fixed geometric structure, which is not reasonable for handling the object with significant appearance variations. For example, when the athlete in Fig.~\ref{fig:def_feat} runs, his body parts rapidly change appearances and locations. The standard convolutional features cannot adapt to such variation well. On the contrary, the deformation convolution roughly estimates how the visual parts of the object will move by $\mathcal{F}_{deform}$ and then resamples the feature maps by bilinear interpolation $\mathcal{F}_{sample}$. It estimates the feature maps after object deformation and thus the approximated feature maps serve as the deformable features, which recovers the deformed object features into {the standard} ones. Since the deformation is simply modeled by the offsets of the sampling grids, it supports free-form deformation (e.g., rigid or non-rigid deformation, pose variation of human body and rotation) instead of affine transformation in STN~\cite{jaderberg2015spatial}. Therefore, it is useful for tracking objects with not only non-rigid deformation but also in-plane rotation.
Besides, the performance of the standard CNNs is degraded because the training samples collected online are limited and similar. Due to the online estimated offsets, the deformable convolution can adaptively model the unseen deformation in an online manner.

\begin{figure}
	\centering
	
	\includegraphics[width=0.45\textwidth]{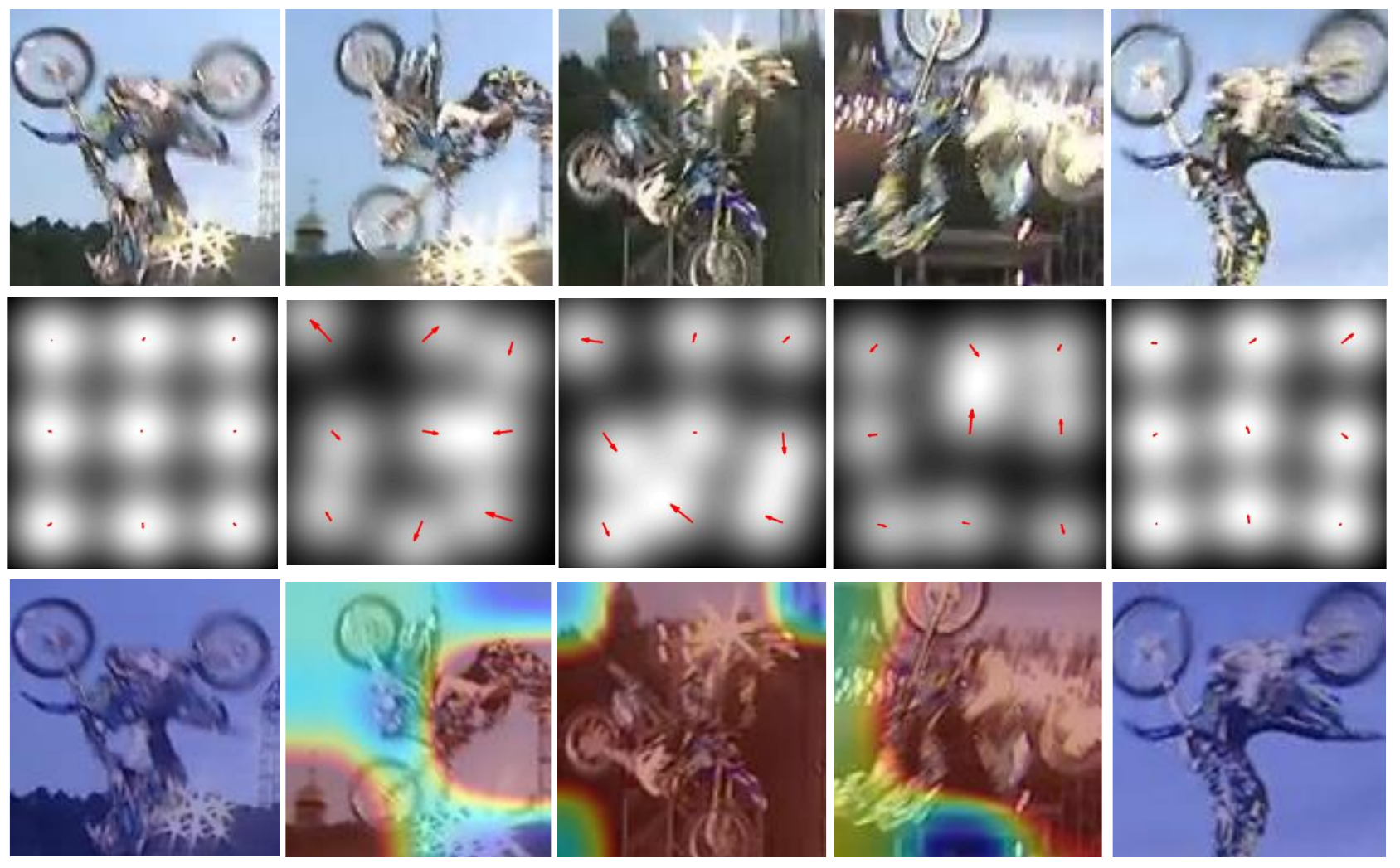}
	
	\caption{{In the first row, the tracked image patches from
		{frames \#2, \#21, \#31, \#79, and \#86 of the \textit{MotorRolling} sequence are shown. In} the second row, the deformation offsets are visualized. The bright regions and the red arrows indicate how the deformation offsets affect the feature extraction. In the third row, the differences between the {standard convolutional feature maps and the deformed convolutional feature maps} are highlighted. We select the feature maps from the $30^{th}$ channel to compute the difference, which is mapped back to the corresponding image patch for visualization. Warmer colors indicate larger differences.}}
	
	\label{fig:deform1}
	
\end{figure}

{In Fig.~\ref{fig:deform1}, we illustrate an example of the effect of the deformable convolutional features during online tracking.
As shown, we select some exemplar frames in which the target rotates and changes its pose from frames \#2 to \#79. We observe that the deformable convolution module adapts to the pose variation by extracting features according to the learned deformation offsets.
Note that in frame \#86, when the target recovers back to the original pose, the deformable convolution features also recover to the standard convolutional features.
In summary, the deformable convolutional features are adaptive when capturing the appearance variations of the deformable object online.}

\subsection{Gated Fusion}
\label{sec:gate}



In practice, the deformable convolution layer may not perform well and its output feature maps will be erroneous. So, it needs to be compensated for by the standard convolutional features. Here, we introduce a gating module to control the fusion of both features, as shown in Fig.~\ref{fig:framework}. The gating mechanism is first proposed in Long-Short Term Memory (LSTM) cells~\cite{hochreiter1997long} for regulating the information flow through the network. 

In our framework, we introduce a soft-gate to adaptively fuse the deformable convolutional features and standard convolutional features. The motivation of applying the gating module is to enhance the features fusion between the deformable convolutional features and the standard convolutional features. As {mentioned} in Sec.~\ref{sec:feat}, when tracking {an object} with severe appearance variations, the {standard} convolutional features may fail due to the unseen appearance of the object, while the deformable convolutional features can to some extent adapt to the deformation. However, the deformable convolution also has limitations. As we observe, in ordinary scenarios, the {standard} convolutional features demonstrate more robust performance in tracking objects with few appearance variations. In addition, in some extreme situations, such as significant illumination changes or severe occlusion, the deformation convolution cannot accurately estimate the deformation offsets, {and} it may {produce in higher errors than the standard} convolutional features. Thus, we do not only switch between these features, but also combine them adaptively as well.

The output of {the} gating module is learned by another branch of layers, $\mathcal{F}_{gate}$, followed by a sigmoid activation function, $\mathcal{F}_{\sigma}$. The sigmoid layer constrains the gating output to be within $[0,1]$, which serves as a weight that balances which type of features should be dominant. Thus, the output of the gating module is:
\begin{align}
{\sigma} = \mathcal{F}_\sigma(\mathcal{F}_{gate}(\textbf{X})).
\end{align}

To better control the fusion, we set the dimension of the output as $\sigma \in \mathbb{R}^{H\times W}$, which is the same as the spatial dimension of the {standard} convolutional features and the deformable convolutional features.
Given the deformable convolutional features $\textbf{X}^\prime$ and the {standard} convolutional features $\textbf{X}$, the fusion process is computed as:
\begin{align}
\textbf{Y} = \textbf{X}^\prime \odot \sigma + \textbf{X} \odot (1-\sigma),
\end{align}
\noindent where $\odot$ indicates the element-wise multiplication, and $\textbf{Y}$ refers to the fused features. When an object shows appearance variations, the gating module should compute a large $\sigma$. Thus, the deformable convolutional features will be dominant in the fused features. When the deformable convolution cannot achieve good performance, the gating module will output a small $\sigma$ value.

\begin{figure}
	\centering
	\begin{tabular}{c@{}c@{}c}
		\includegraphics[width=0.45\textwidth,height=6cm]{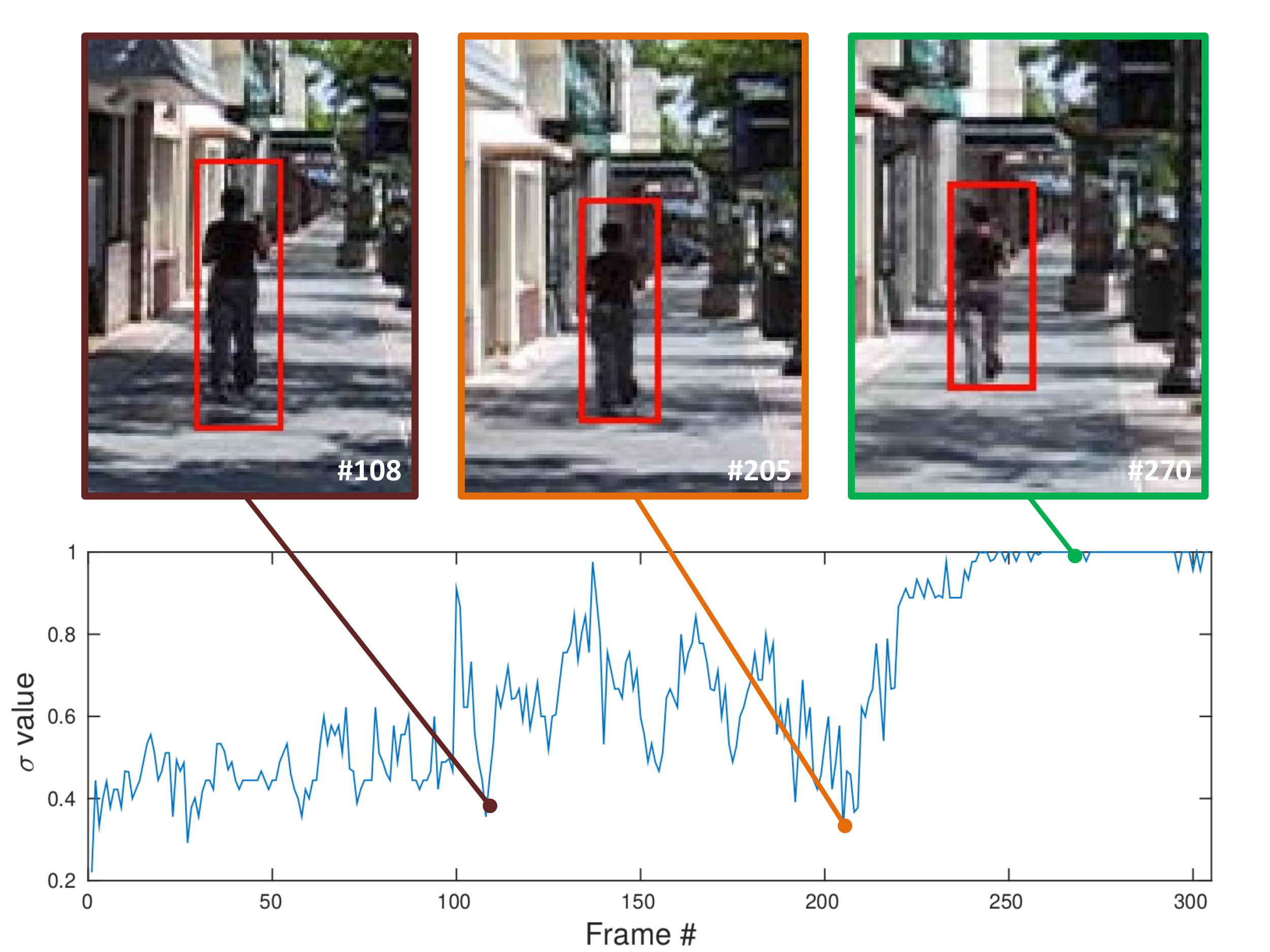} \\
		{\footnotesize{(a)}}\\
		\includegraphics[width=0.4\textwidth]{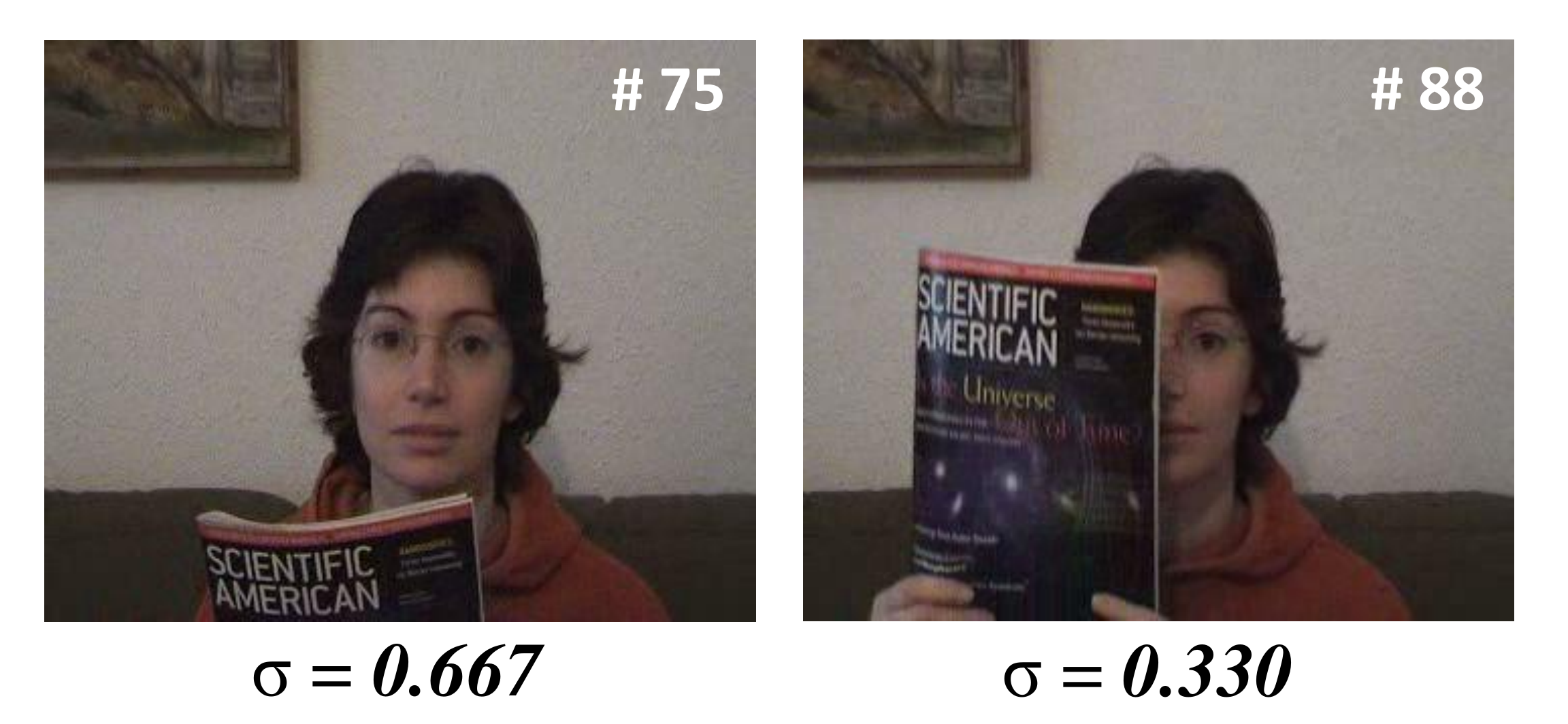}\\
		{\footnotesize{(b)}}
		
	\end{tabular}
	
	\caption{We demonstrate how the online learned gating output (i.e., the $\sigma$ value) works in tracking. (a) shows how the $\sigma$ value changes when tracking video \textit{Human9}. We highlight three keyframes and their corresponding $\sigma$ values. (b) shows the $\sigma$ values of the two frames from video \textit{FaceOcc1}.}
	
	\label{fig:sigma}
	
\end{figure}

In Fig.~\ref{fig:sigma}, we illustrate two examples of how the gating module works in online tracking. The gating module is offline trained with diversified samples and then performs online updating. To simplify the analysis, we compute the average value of $\sigma$. If the $\sigma$ value is equal to $0$, it means that there is no flow passing through the gate in the branch of the deformable convolution module and the fused features will be completely from the standard features. If the $\sigma$ value is equal to $1$, the full flow of the deformable convolution will be enabled and the standard features will not be used.

Fig.~\ref{fig:sigma}(a) illustrates how the gating output (i.e., the $\sigma$ value) is online learned in the $Human9$ sequence, in which a person walking in the street with appearance deformation, scaling and illumination variations. We highlight three keyframes and their corresponding $\sigma$ values.
As we observe, when the person walks into the shadow in {frames} \#108 and \#209, $\sigma$ drops to below $0.4$ as the illumination sudden changes, making it difficult for the deformable features to model the target appearance accurately. Thus, the standard convolutional features take charge. In the end, $\sigma$ increases to 1 as the person walks out of the shadow and its appearance is well observed again. Thus, the deformable convolutional features are used. Note that in the beginning of tracking, the gating output is relatively low because the gating module is {under} initialization by collecting training samples. Fig.~\ref{fig:sigma}(b) reflects that the deformable convolutional features may fail in occlusions. Thus, the standard convolutional features will take the major role in the fusion.
As observed, when the face is occluded, the $\sigma$ value significantly drops from $0.667$ to $0.330$.


In summary, our proposed framework mainly consists of two modules: the deformable convolution module and the gating module. The deformable convolution module generates the deformable features to handle the object appearance variations, while the gating module adaptively fuses the deformable convolutional features and {standard} ones in an online manner.




\subsection{Tracking}
As depicted in Fig.~\ref{fig:framework}, our approach leverages the pretrained VGG-M~\cite{Chatfield14} as the front end to extract standard features in the first stage. In the second stage, the extracted {standard} feature passes through {some} parallel branches for gated feature fusion. In the final stage, the fused features are transferred to fully connected layers, which serve as a non-linear classifier.

\noindent\textbf{Model training.} We train our model using positive and negative samples from the training data offline. We prepare the training data following~\cite{nam2016learning}. For offline training, 50 positive and 200 negative samples are collected from {each} frame,
where positive and negative examples have $\geq 0.7$ and $\leq 0.5$
IoU overlap ratios with ground-truth bounding boxes, respectively. To train the deformable convolution layers and the gating module, we follow three steps: 1) training the network without deformable convolution and gating; 2) training the network with the deformable convolution only; 3) training the network with both modules. Hence, in steps 1 and 2, we actually fine-tune the standard convolutional feature and the deformable convolutional feature, respectively. In step 3, we train the gating module to make sure the tracker can adaptively switch between both modules. All of them are trained in an end-to-end manner.

\noindent\textbf{Model initialization.} With the trained model, we fine-tune the network using the samples as the candidate proposals from the first frame of the input sequence.
The front end of the network is frozen and only the gating feature fusion and the fully connected layers are updated. To estimate the scale of the object, following the practice of \cite{nam2016learning}, we train a bounding box regression using a simple linear regression model to predict the precise target location using $conv3$ features of the samples near the target location at the first frame.


\noindent\textbf{Model update.} We incrementally update the tracker online. Around the estimated position, we generate multiple samples and assign them with binary labels according to the intersection-over-union ratios with the estimated bounding box. The optimal target state is the drawn sample with the highest positive score. In addition, we adopt hard negative mining in mini-batch sampling. The hard negative samples are selected from the classification confidence of samples, which are evaluated in our network. In particular, we select the negative samples with top confidence scores as the hard negative samples, which are used for updating the model. 

	\section{Experiments}
	\label{sec:exp}
	
	In this section, we introduce the implementation details of our proposed model and analyze the effects of the modules in the network. We refer to our tracker as \textbf{GDT} (\textbf{G}ated-fusion \textbf{D}eformable object \textbf{T}racker) and we compare GDT with state-of-the-art trackers in the benchmark datasets: Deform-SOT~\cite{du2016online}, OTB-2013~\cite{wu2013online}, OTB-2015~\cite{wu2015object}, VOT-2016~\cite{kristan2013visual}, and VOT-2017~\cite{VOT_TPAMI_2017} for performance evaluation.

	\subsection{Implementation Details and Experimental Setup}
	
	\noindent \textbf{Network architecture.}
	First, our feature extraction network is based on the first three convolutional layers from the VGG-M model~\cite{Chatfield14}.
	Second, for the deformable convolution module, it consists of a convolutional layer and a fully connected layer, which generates the deformation offsets {of} $3\times3 \times 2$. Then, the feature maps are reconstructed according to the offsets by a bilinear sampler. For the gating module, it is produced by two consecutive fully connected layers followed by a sigmoid activation and outputs a $3\times3$ gating values. {Finally}, {the classifier consists of three consecutive fully connected layers, with a softmax loss as the objective function.}
	

	\noindent\textbf{Implementation details.}
	For offline training the network, we apply {the} Sochastic Gradient Descent (SGD) solver to run for $200K$ iterations. The learning rate of the feature extraction convolutional layers is set as $10^{-4}$, and that of the fully connected layers is set as $10^{-3}$. For the evaluation in OTB, the training data is collected from labeled sequences of VOT challenges excluding the sequences from OTB-100. For the evaluation in VOT-2016 and VOT-2017, we use OTB data for training excluding the sequences in {the} test sets. {For the test in Deform-SOT, our model is trained using the data from OTB and VOT excluding the test sequences used in Deform-SOT.}
    At the initialization stage, we train the feature fusion and the fully connected layers for 30 iterations with {a} learning rate $0.0003$ except for the last layer, which is $0.001$. For online update, the feature extraction layers are frozen and the rest of the network is fine-tuned based on online collected samples.
    We assign a learning rate of $0.0005$ for the feature fusion module. The momentum and weight decay are always set to $0.9$ and $0.0005$, respectively. Each mini-batch consists of 32 positives and 96 hard negatives selected out of 1024 negative examples.
	Our proposed tracker GDT runs on a PC with an i7 3.6GHz CPU and a NVIDIA Geforce 1080Ti GPU with the MatConvNet toolbox~\cite{Vedaldi2015MatConvNet}. In tracking, the initial training time is around 30 seconds on average and the running time of tracking is $1.33$ FPS. The tracker is updated every $10$ frames online and the updating time is around $3$ seconds.

	\noindent\textbf{Evaluation metrics.} We follow the standard evaluation approaches. In the Deform-SOT, OTB-2013 and OTB-2015 datasets, we use
	the one-pass evaluation (OPE) with precision and success
	plots metrics. The precision metric measures the frame locations rate within a certain threshold distance from {the} groundtruth locations. The threshold distance is set to $20$ pixels.
	The success plot metric is set to measure the overlap ratio between the predicted bounding boxes and the groundtruth.
	\ryn{For the VOT-2016~\cite{kristan2013visual} and VOT-2017~\cite{VOT_TPAMI_2017} datasets, we conduct comprehensive experiments including reset-based experiments and no-reset (unsupervised) experiments. The reset-based experiments are measured in terms of Expected Average Overlap (EAO), Accuracy and Robustness. Specifically, robustness is computed as $\exp(-S\cdot F)$, where $F$ refers to the tracking failure rate and $S$ is set as $100$, according to \cite{Cehovin2016Visual}.
		The unsupervised experiments are evaluated in terms of Average Overlap (AO).}
	Note that in the tables included in the following subsections, \ryn{we mark the best, second best and third best results in \textit{red/blue/green} colors, respectively}.

	\begin{figure}
		\hspace{-.2in}
		\begin{tabular}{c@{}l@{}c}
			\includegraphics[width=0.26\textwidth]{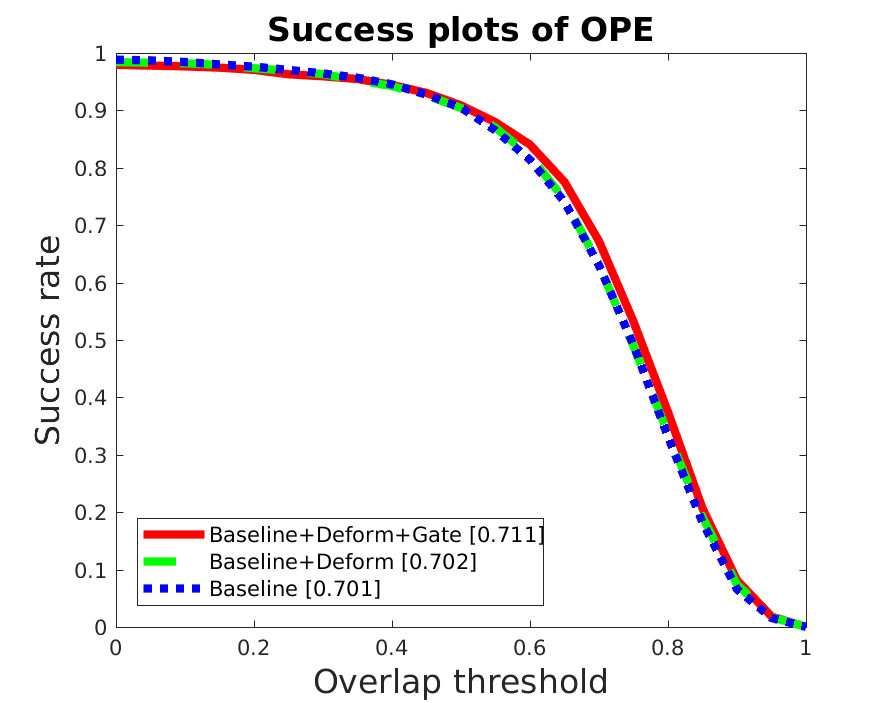} &
			\includegraphics[width=0.26\textwidth]{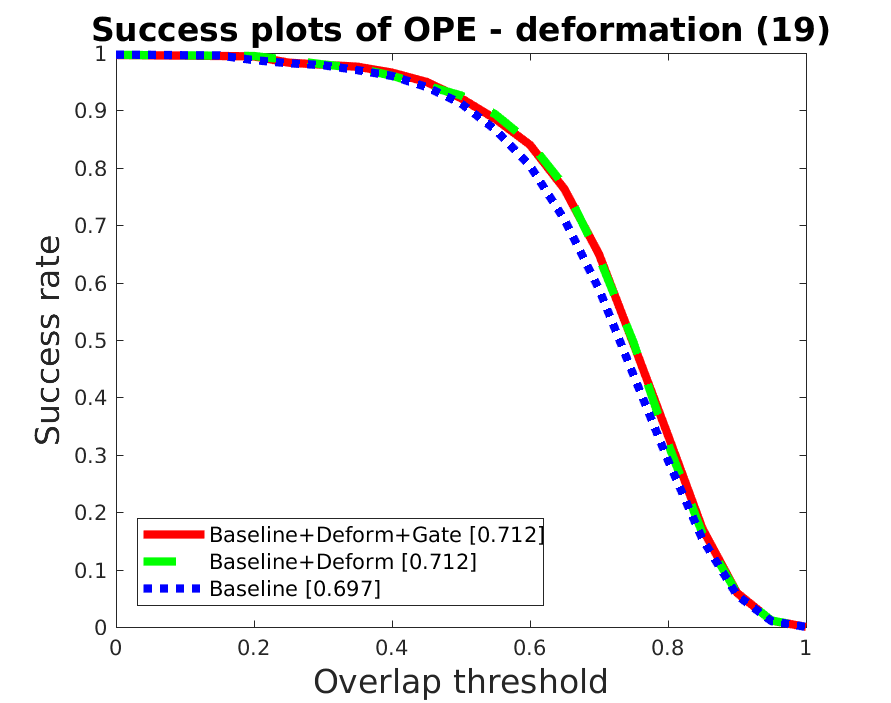}\\
		\end{tabular}
		
		\caption{Success plots on all sequences and the sequences with the deformation attribute in OTB-2013 dataset using the one-pass evaluation. The numbers in the legend indicate the area-under-the-curve success scores. }
		
		\label{fig:ablation}
		
	\end{figure}

	\subsection{Ablation Studies}
	{To validate the effectiveness of each module, we separately train the following three models: network without the two modules (Baseline), network with the deformable convolution module only (Baseline+Deform), and the complete network (Baseline+Deform+Gate).} We compare the performance of these models on the OTB-2013 dataset.	
	In Fig.~\ref{fig:ablation}({left}), we evaluate the overall tracking performance by the success rate metric {on} the entire dataset. In Fig.~\ref{fig:ablation}(right), we measure the tracking performance {on} the videos containing the \textit{deformation} attribute.

	As shown in {Fig.~\ref{fig:ablation}}, the deformable convolutional features help slightly boost the overall success rate from $0.701$ to $0.702$, but {they help improve the capability of tracking deformable objects more significantly} (success rate from $0.697$ to $0.712$).
	This demonstrates that the deformable convolutional features indeed benefit tracking deformable objects. However, the overall tracking capability is not obviously improved. The reason is that, as we {have} mentioned in Sec.~\ref{sec:gate}, the deformable convolutional features are not that robust as the {standard} convolutional features.
	After incorporating the gating module, the overall performance improves more significantly (from $0.702$ to $0.711$), while not degrading the capability of tracking deformable objects (remaining at $0.712$). This result indicates that the gating module supplements the deformable convolutional features and helps improve the general tracking performance. More importantly, the capability of tracking deformable objects is not affected, which means {that} the gating module can adaptively switch the fusion to the deformable convolutional features when tracking deformable objects.
    As illustrated in Fig.~\ref{fig:ablation}(c-f), the deformable convolutional features do not significantly improve or even degrade the tracking performance. However, as we {can} observe, the {inclusion} of the gating module plays an important role to boost the performance.
	
\begin{figure}
		\centering
		
		\includegraphics[width=0.45\textwidth]{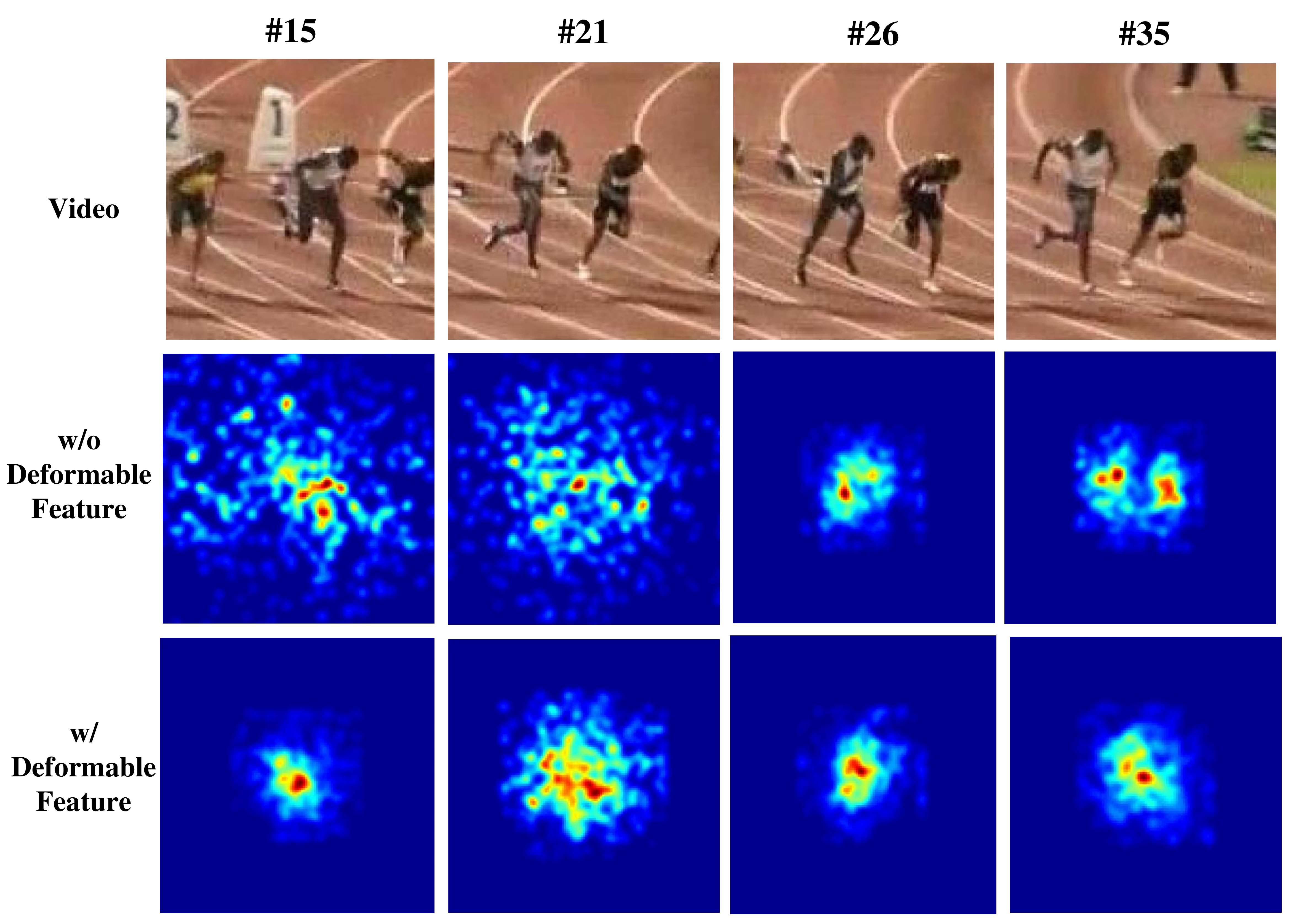}
		
		\caption{A comparison of the predicted probabilities (i.e., confidence scores) from the classifier with and without {the} deformable convolutional features. To visualize this, we densely draw the samples {from} the region of interest and measure their confidence scores.
			Their classification confidence scores with regards to their locations are illustrated by heatmaps. The exemplar images are selected from the \textit{Bolt2} video in the OTB-2015 benchmark.}
		
		\label{fig:def_feat}
		
	\end{figure}

\begin{figure}
		\vspace{-.1in}
		\begin{tabular}{c@{}c@{}c}
			\includegraphics[width=0.25\textwidth]{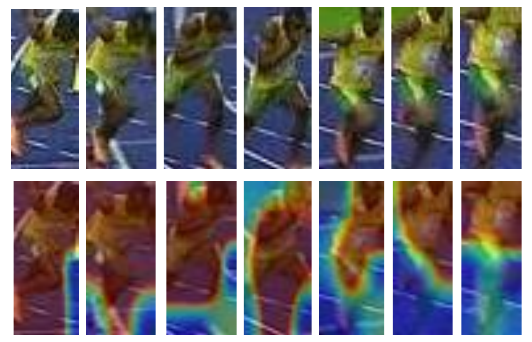} &
			\includegraphics[width=0.25\textwidth,height=2.92cm]{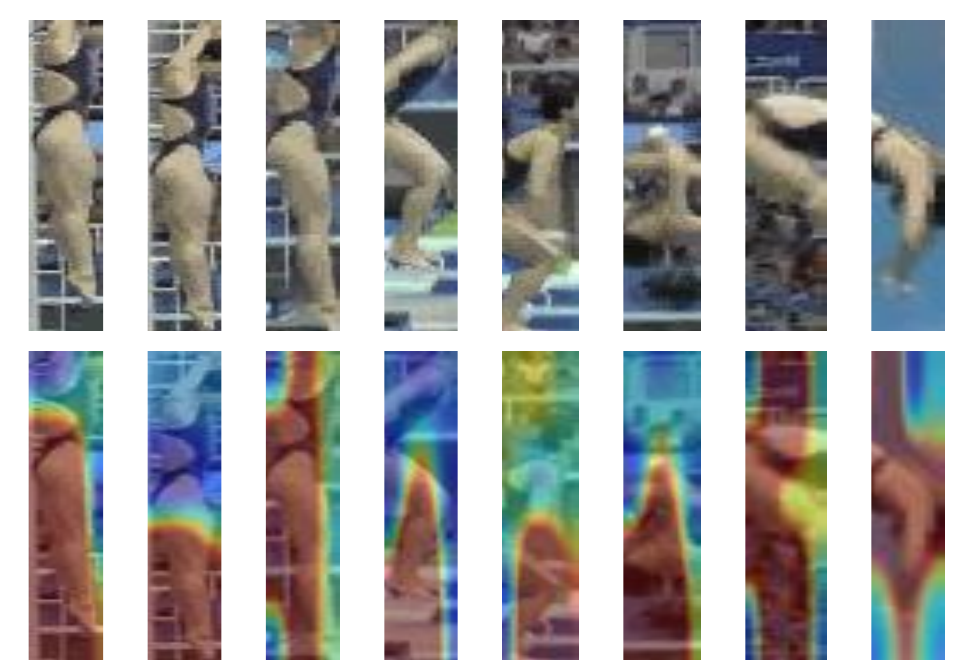}\\
			            \footnotesize (a) \textit{Bolt} & \footnotesize(b) \textit{Diving}\\
		\end{tabular}
		\caption{{In} the first row of (a) and (b), the exemplar tracked image patches from $Bolt$ and $Diving$, respectively, are shown. In the second row, the differences between the standard convolutional feature maps and the deformed convolutional feature maps are highlighted. In (a), patches from frames
			\#3, \#12, \#29, \#41, \#54, \#58, and \#66
			are selected for visualization, while in (b), patches are selected from frames
			\#8, \#16, \#31, \#44, \#47, \#97, \#119, and \#199.}
		
		\label{fig:deform2}
		
	\end{figure}

	{\noindent \textbf{Deformable convolution.} To {demonstrate} the effects of the deformable convolution module, we choose the video sequences \textit{MotorRolling}, \textit{Bolt}, \textit{Bolt2}, and \textit{Diving}, which contain in-plane rotation and deformation, from the OTB-2015 dataset for {testing}.
	{First}, as shown in Fig.~\ref{fig:deform1}, with the pose variation, the deformable convolution module adaptively focuses on the rotating body of the target.
	{Second}, in Fig.~\ref{fig:def_feat}, we densely draw image patches from the frames \#15, \#21, \#26, \#35 in the \textit{Bolt2} video and pass them to the network to compute their classification confidence scores. This figure {shows} the locations of the drawn samples and their corresponding confidence scores.
	As observed, without deformable features, the classifier is uncertain around the target region when the athlete runs with appearance variations. In frames \#15 and \#21, the high confidence samples disperse, which indicates {that} the target pose affects the performance of the tracker. In frame \#35, the confident samples are gathered in two clusters, which means {that} the tracker is confused by the target object and the neighboring object.
	In contrast, the high confidence area of the classifier with the deformable convolutional features is more concentrated on the target object, even though the target object exhibits some unseen poses.
	{Third}, in Fig. \ref{fig:deform2}, the differences between the deformable features and the {standard} features are visualized. As observed, the major differences correspond to the body parts with significant deformation. {This toy experiment reflects} the adaptive effects of the deformable convolution module during online tracking.}

	\subsection{Quantitative Evaluation}
	
	{\noindent\textbf{Deform-SOT Dataset.} This dataset contains 50 challenging video sequences \cite{du2016online} that focus on tracking totally deformable targets in unconstrained environments. The annotated attributes include scale change, severe occlusion, abnormal movement, illumination variation and background clutter. Following the experiment setup and the results provided in \cite{du2016online}, we compare with 8 part-based trackers: Structure-aware hypergraph based tracker (SAT~\cite{du2016online}), Dynamic graph tracker (DGT \cite{Zhaowei2014Robust}), Temporally coherent part-based tracker (TCP \cite{Li2015Online}), Local and global tracker (LGT \cite{Luka2013Robust}), Superpixel tracker (SPT \cite{Shu2011Superpixel}), Adaptive structural local sparse appearance model (ASLA \cite{Xu2012Visual}), Latent structural learning tracker (LSL \cite{Yao2013Part}), Sparsity-based collaborative model (SCM \cite{Wei2012Robust}). We evaluate our GDT along with these trackers on 50 video sequences using the one-pass evaluation based on {two} metrics: distance precision and overlap success.}
	
\begin{figure*}
		\begin{tabular}{c@{}c@{}c@{}c}
			
			\includegraphics[width=0.25\textwidth,height=3.5cm]{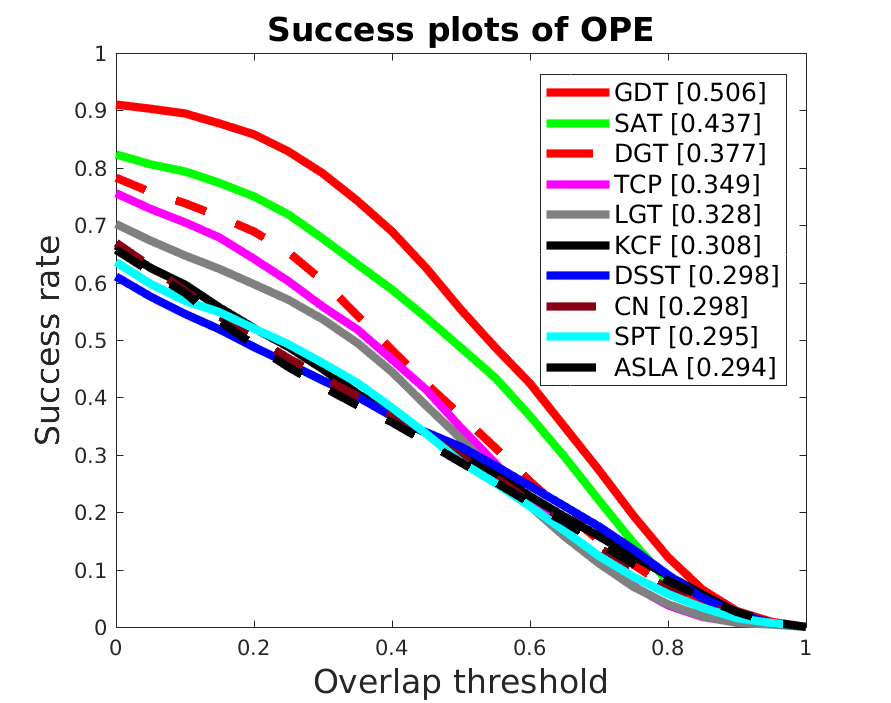}&
			\includegraphics[width=0.25\textwidth,height=3.5cm]{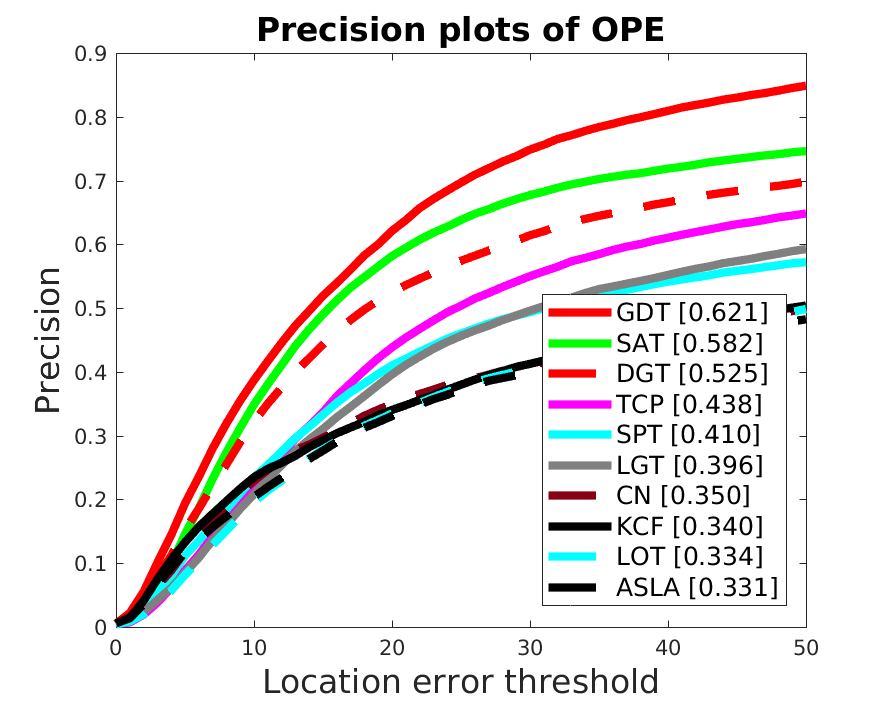}&
			\includegraphics[width=0.25\textwidth,height=3.5cm]{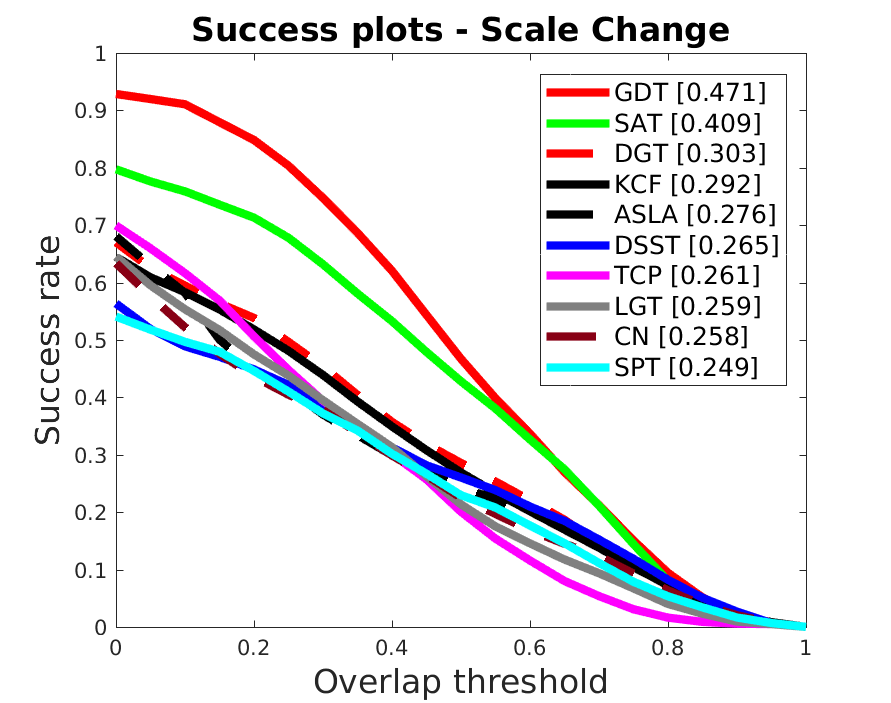}&
			\includegraphics[width=0.25\textwidth,height=3.5cm]{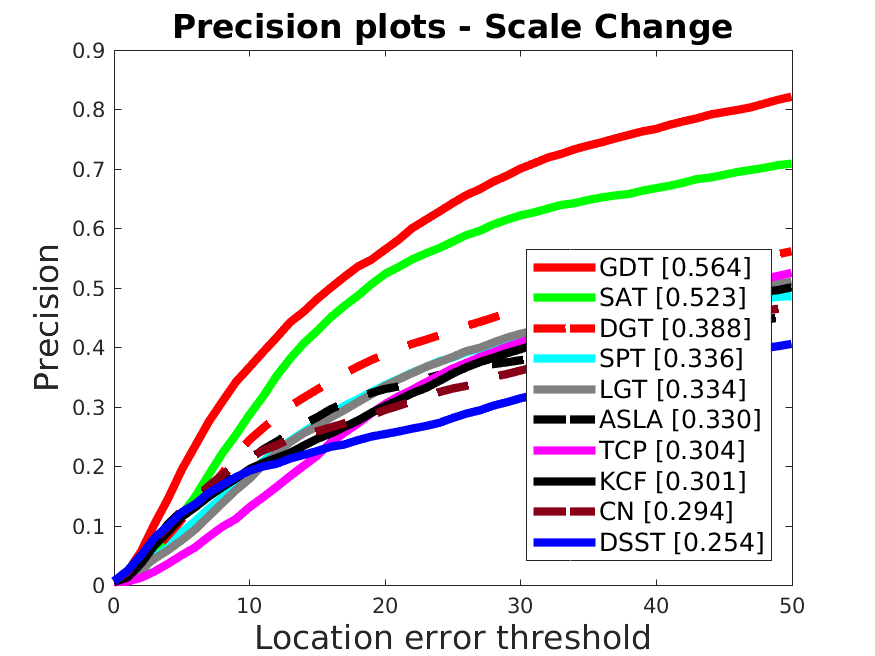}\\
			\includegraphics[width=0.25\textwidth,height=3.5cm]{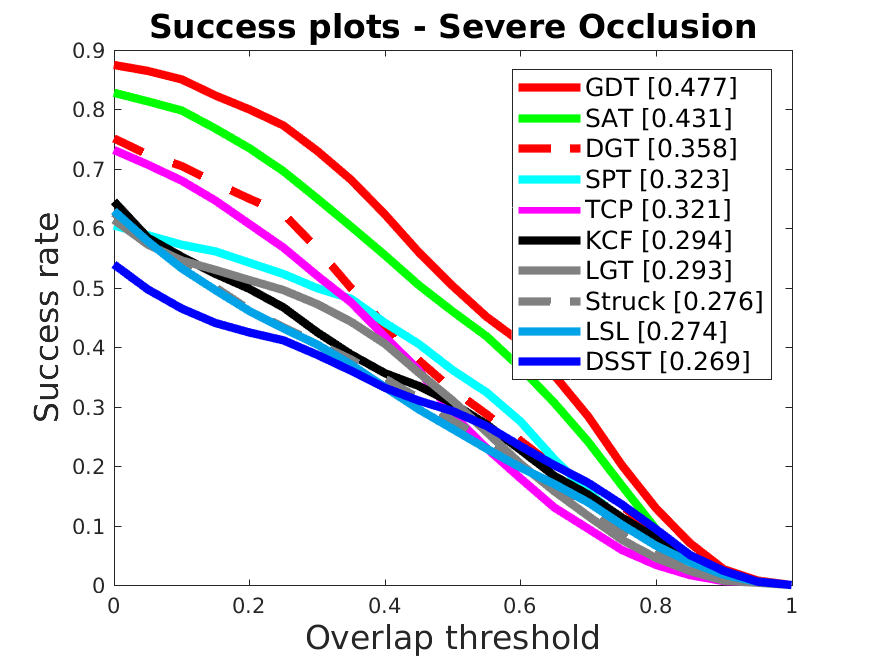}&
			\includegraphics[width=0.25\textwidth,height=3.5cm]{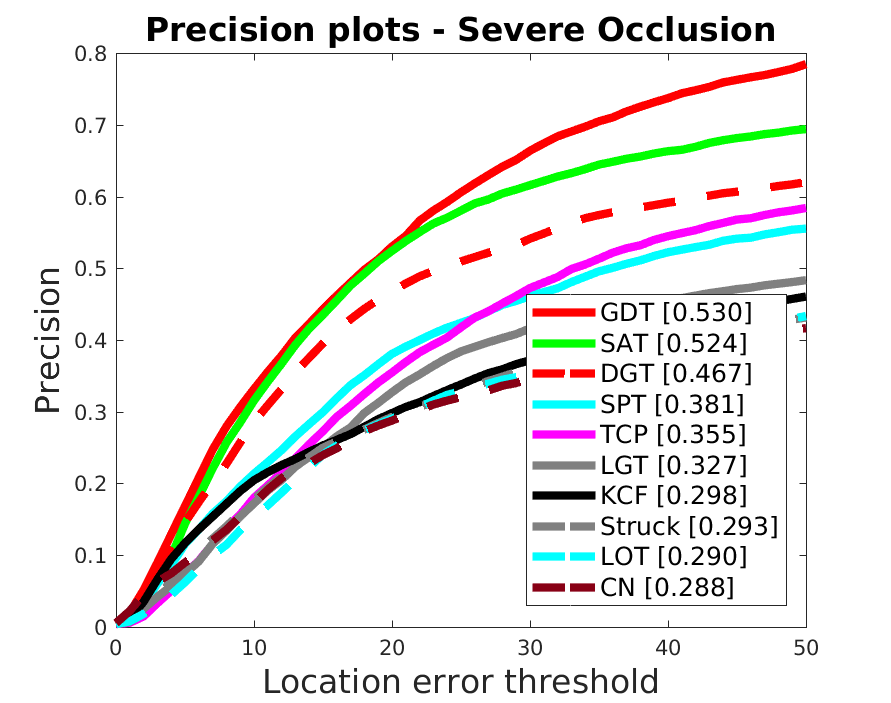}&
			\includegraphics[width=0.25\textwidth,height=3.5cm]{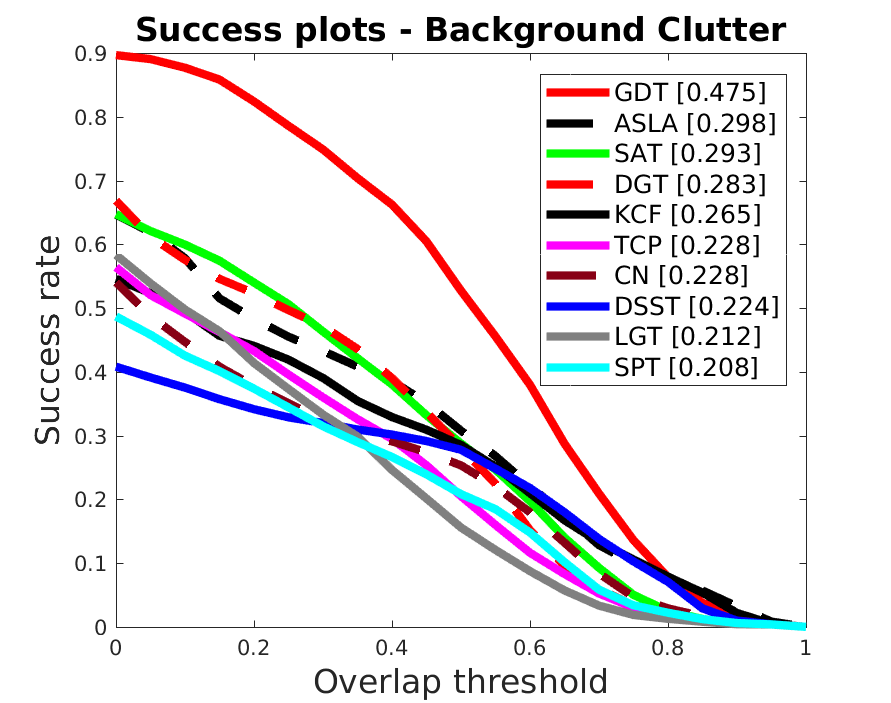}&
			\includegraphics[width=0.25\textwidth,height=3.5cm]{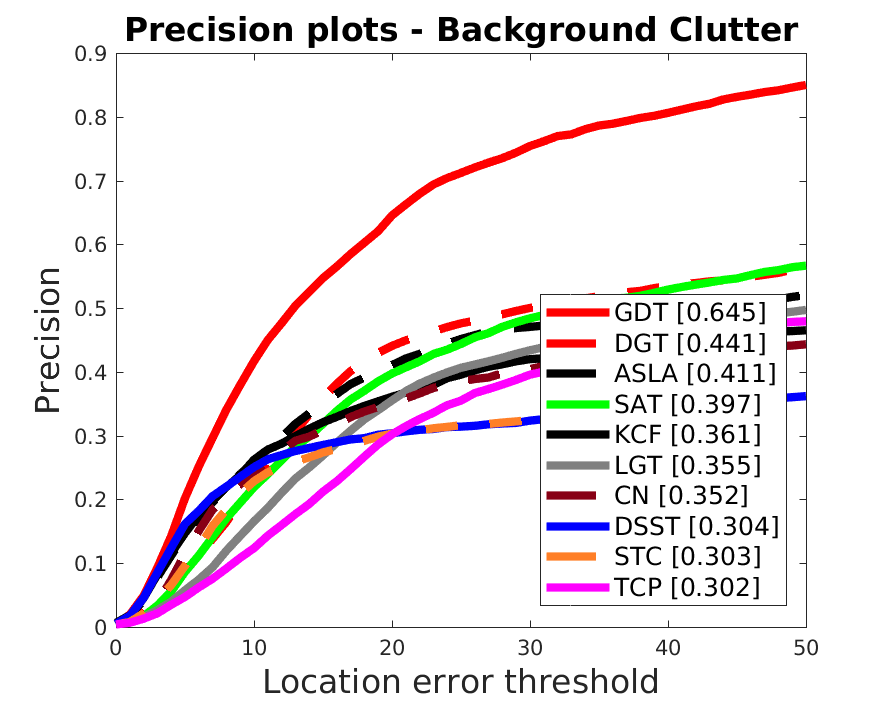}\\
			\includegraphics[width=0.25\textwidth,height=3.5cm]{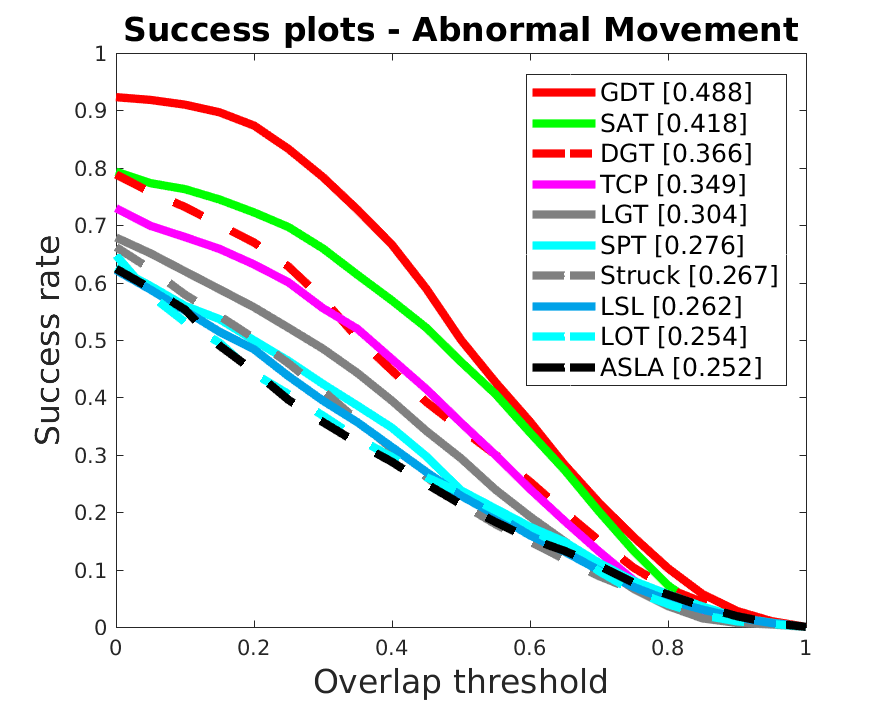}&
			\includegraphics[width=0.25\textwidth,height=3.5cm]{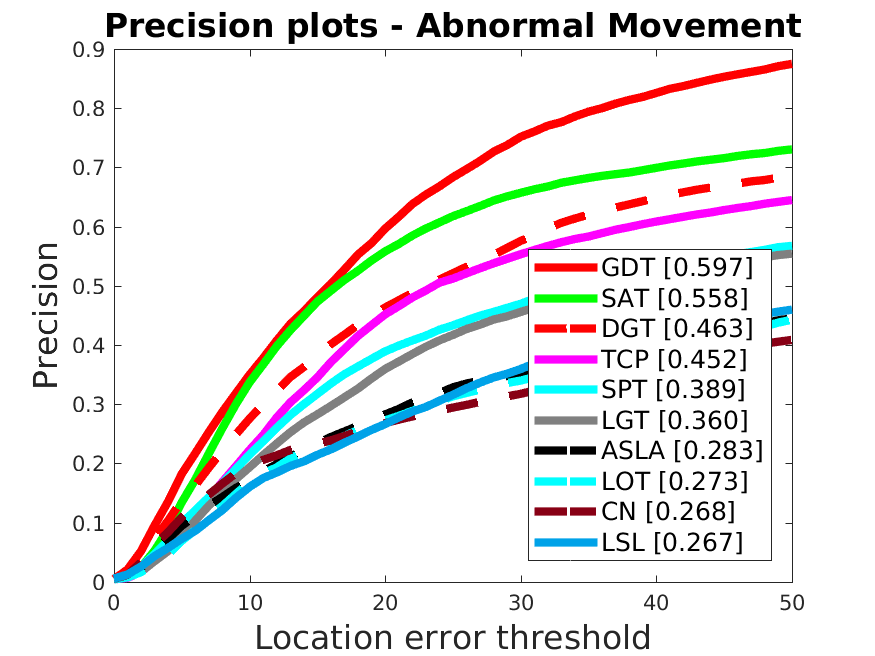}&
			\includegraphics[width=0.25\textwidth,height=3.5cm]{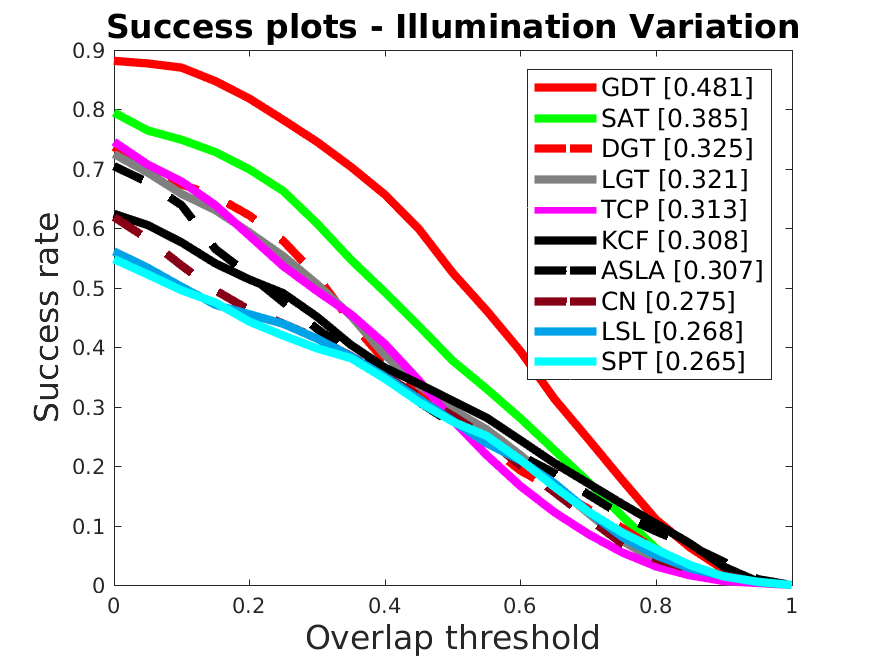}&
			\includegraphics[width=0.25\textwidth,height=3.5cm]{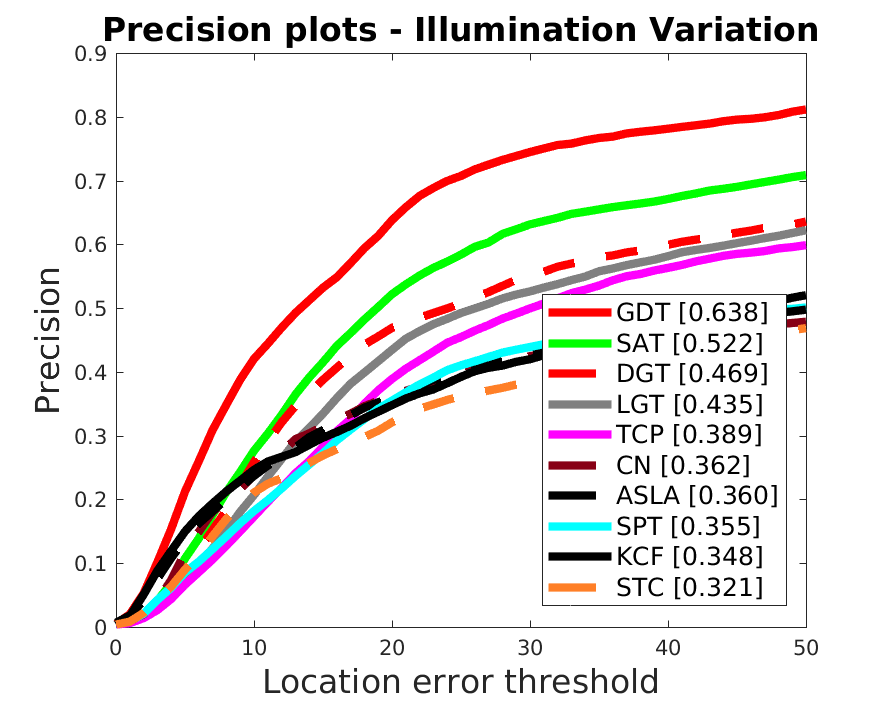}			
		\end{tabular}
		\caption{The top-left first two diagrams show the overall performances on the complete Deform-SOT dataset. The other {diagrams} show overlap success plots and precision plots over sequences annotated with \textit{scale change, severe occlusion, background clutters, abnormal movements}, and \textit{illumination variations} in the dataset.}
		
		\label{fig:deform}
	\end{figure*}

	{In Fig.~\ref{fig:deform}, we show the results of the comparison with the state-of-the-art deformable object trackers. The top-left first two {diagrams} {show} the overall performances of the overlap threshold and the location error, in which our approach significantly outperforms other methods. In the remaining diagrams, we show the results on videos annotated as scale change, severe occlusions, background clutter, abnormal movement, and illumination variation. Again, our approach performs better than all the other part-based trackers.}
	
	\noindent\textbf{OTB-2013 Dataset.} We compare GDT with 29 trackers from the OTB-2013 benchmark~\cite{wu2013online} and other 27 state-of-the-art trackers including TCNN~\cite{Nam2016Modeling}, EBT~\cite{Zhu2016Beyond}, DSST~\cite{danelljan2014accurate}, KCF~\cite{henriques2015high}, MEEM~\cite{zhang2014meem}, LCT~\cite{ma2015long}, MUSTer~\cite{hong2015multi}, HCFT~\cite{ma2015hierarchical}, FCNT~\cite{wang2015visual}, SRDCF~\cite{danelljan2015learning}, CNN-SVM~\cite{hong2015online},	DeepSRDCF~\cite{danelljan2015convolutional}, Staple~\cite{bertinetto2016staple}, SRDCFdecon~\cite{danelljan2016adaptive}, CCOT~\cite{danelljan2016beyond}, SiamFC~\cite{bertinetto2016fully}, MDNet~\cite{nam2016learning}, ADNet~\cite{yoo2017action}, ECO~\cite{danelljan2017eco}, MCPF~\cite{zhang2017multi}, and CREST~\cite{song2017crest}. We evaluate all the trackers on 50 video sequences using the one-pass evaluation with distance precision and overlap success metrics.
	
\begin{figure}
		\centering
		\begin{tabular}{c@{}c@{}c@{}c@{}c}
			
			\includegraphics[width=0.26\textwidth,height=3.5cm]{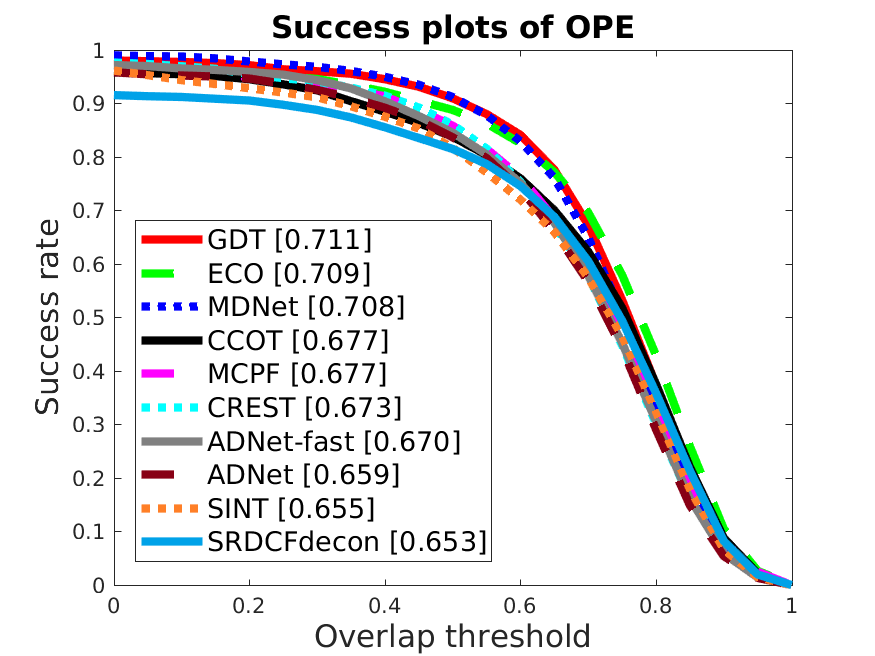}&
			\includegraphics[width=0.25\textwidth,height=3.5cm]{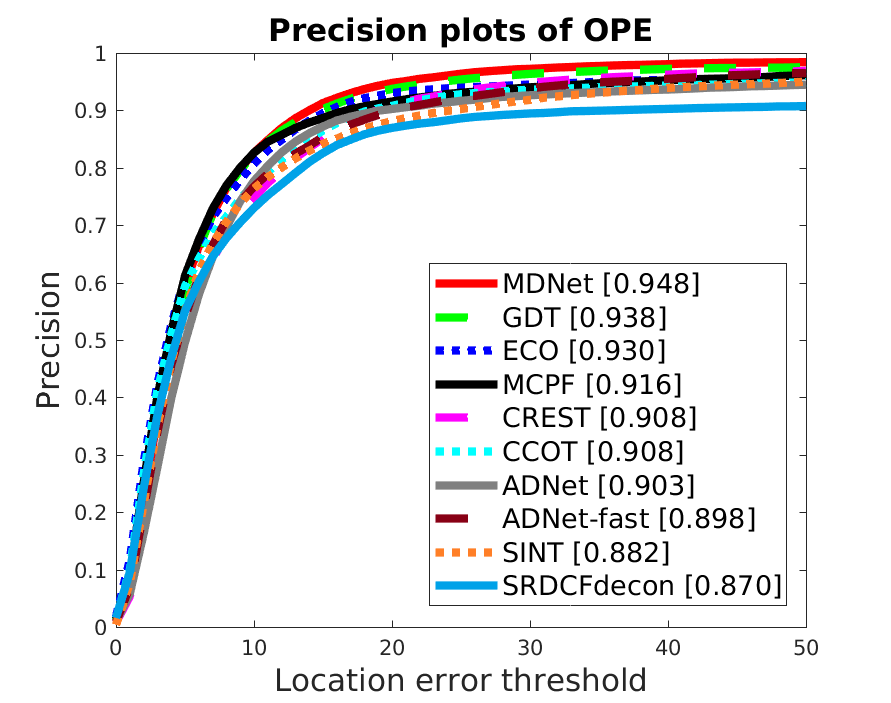}
			
		\end{tabular}
		
		\caption{Precision and success plots on OTB-2013 using one-pass evaluation.}
		
		\label{fig:2013}
		
	\end{figure}

	Fig.~\ref{fig:2013} shows the quantitative results from {the} compared trackers. For presentation clarity, we only show the top 10 trackers. The numbers listed in the legend indicate the AUC overlap success and 20 pixel distance precision scores. Overall, our GDT performs favorably against state-of-art trackers in both distance precision and overlap success. GDT achieves the top success rate ($0.711$) {compared to} other methods including ECO and MDNet. For the distance precision, GDT performs better than other trackers except for MDNet, {which} is similar to our baseline. It means that the deformable convolution slightly decreases the robustness of the model.
	
\begin{figure}
		\vspace{-0.2in}
		\begin{tabular}{c@{}c@{}c}
			
			\includegraphics[width=0.25\textwidth,height=3.5cm]{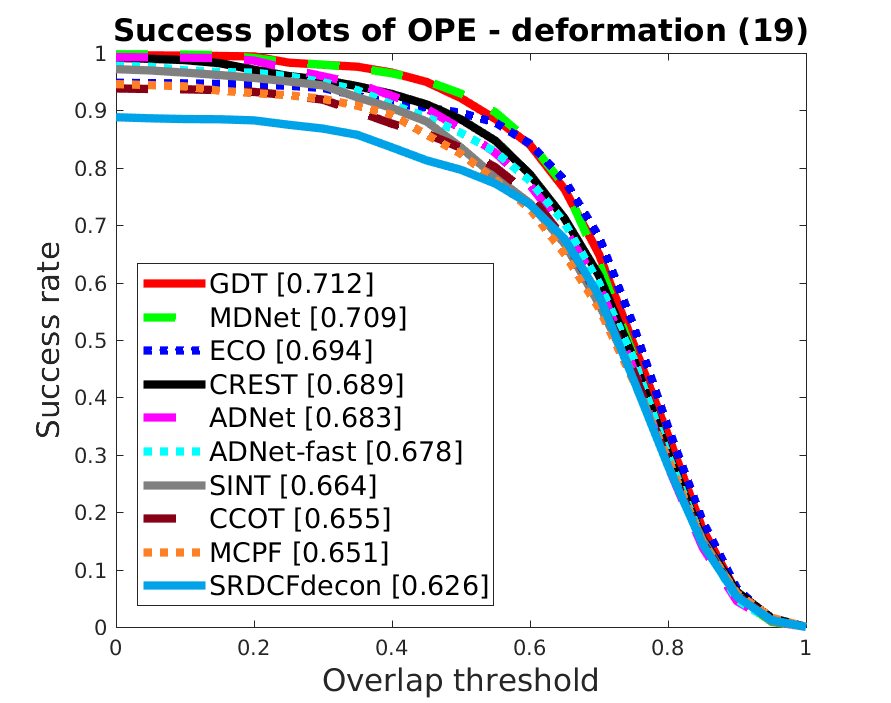}&
			\includegraphics[width=0.25\textwidth,height=3.5cm]{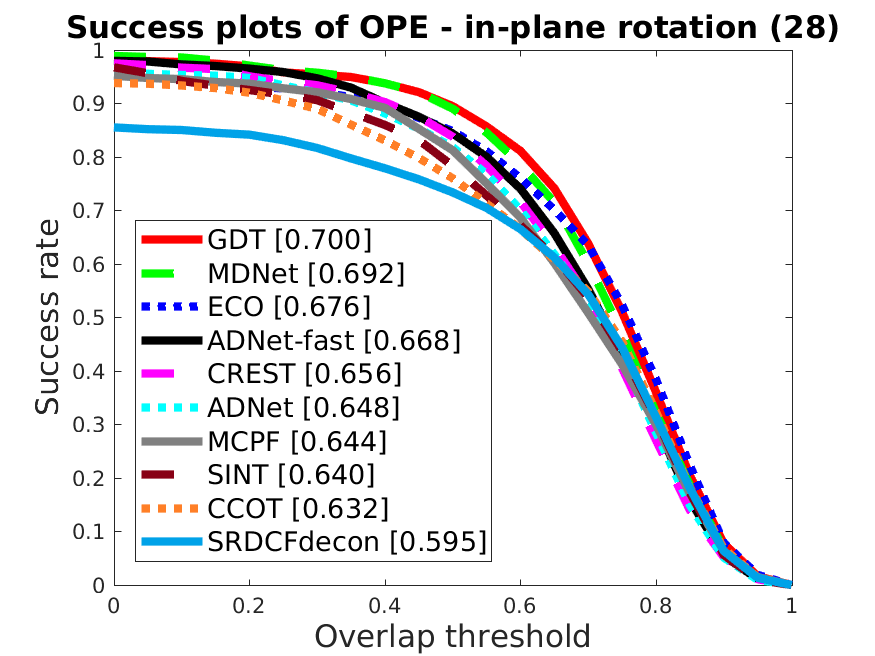}\\
			\includegraphics[width=0.25\textwidth,height=3.5cm]{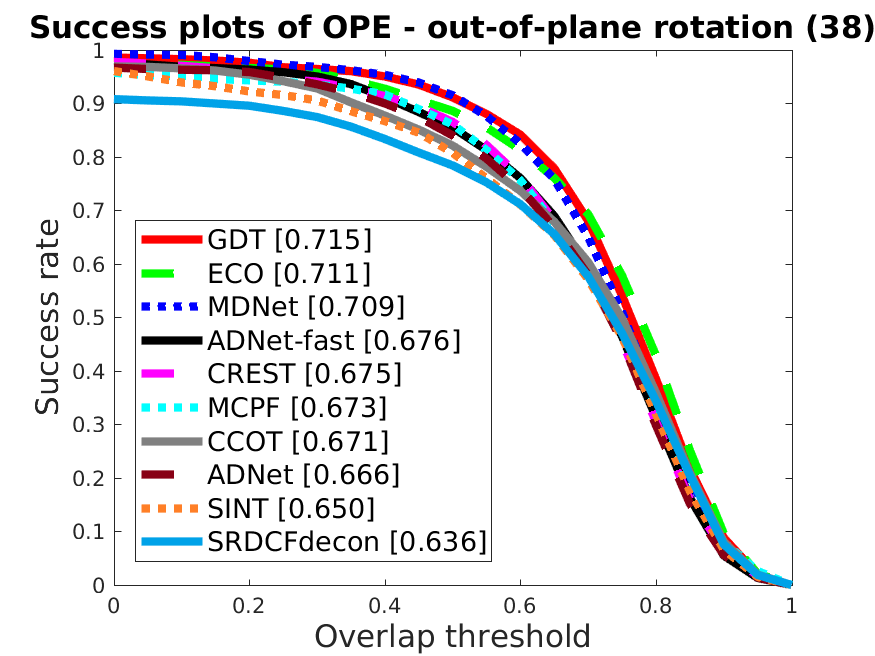}&
			\includegraphics[width=0.25\textwidth,height=3.5cm]{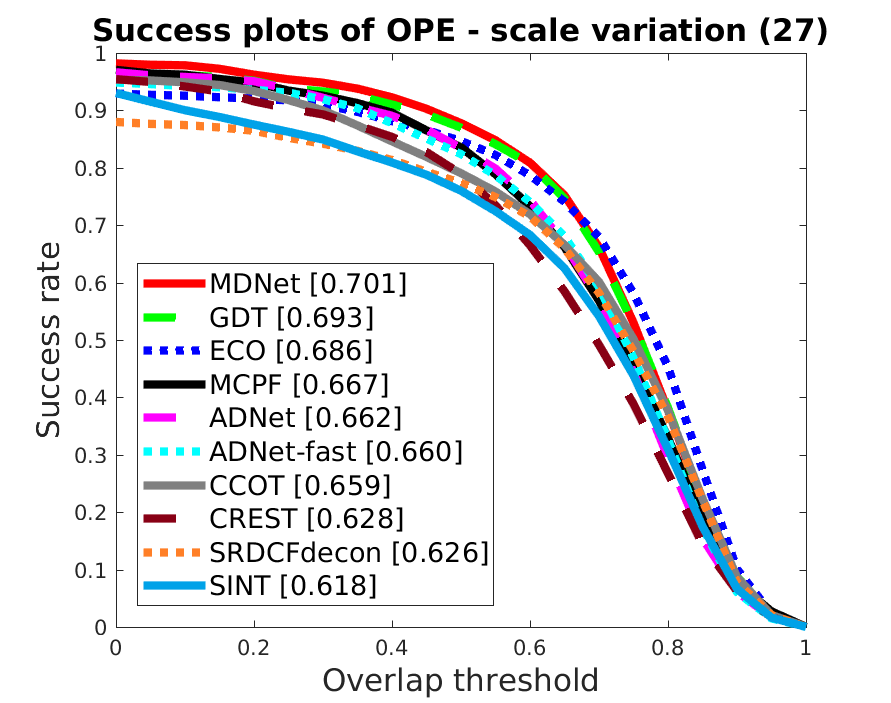}		
		\end{tabular}
		\caption{Overlap success plots over tracking challenges with deformation, in-plane rotation, out-of-plane rotation, and scale variations {on} OTB-2013.}
		
		\label{fig:attr}
	\end{figure}

	In addition, Fig.~\ref{fig:attr} compares the performance of selected six video attributes including \textit{deformation, in-plane rotation, out-of-plane rotation}, and \textit{illumination variation}, using one-pass evaluation. {As can be seen}  from the results, our tracker GDT generally {outperforms} most of the state-of-the-art trackers.
	In particular, our tracker handles large appearance variations caused by deformation, in-plane and out-of-plane rotations better than all of the other methods. It indicates that the deformable convolution {module} enhances the capability of modeling a variety of appearance variations. {In addition, the gating module blends the standard} convolutional features and the deformable convolutional features, which to some extent can model the unseen appearance variations by approximately recovering them to the normal conditions.
	
	 \begin{figure}
		\centering
		\begin{tabular}{c@{}c@{}c@{}c@{}c}
			
			\includegraphics[width=0.25\textwidth,height=3.5cm]{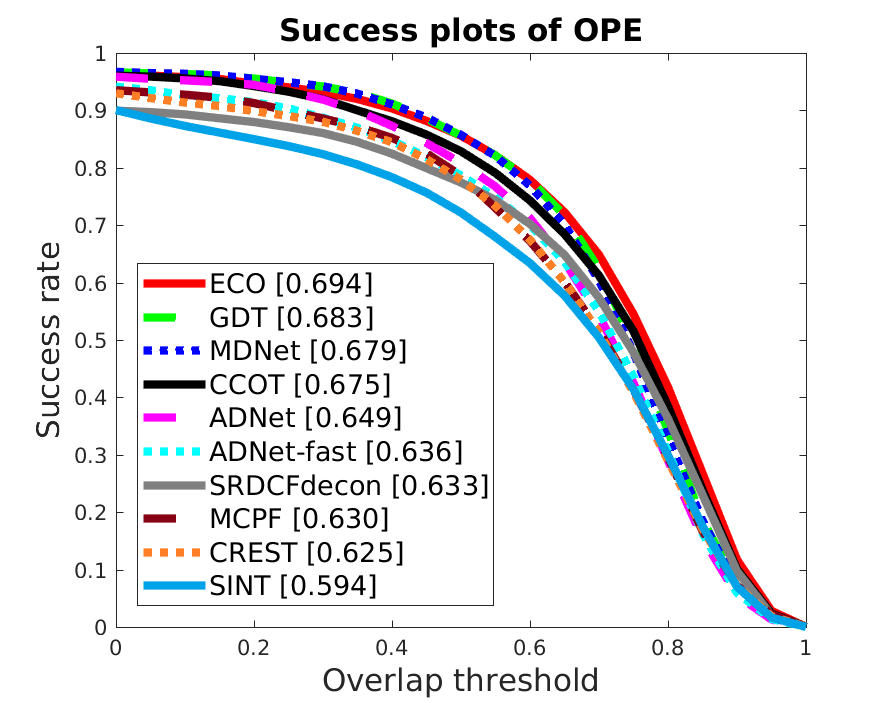} &
			\includegraphics[width=0.25\textwidth,height=3.5cm]{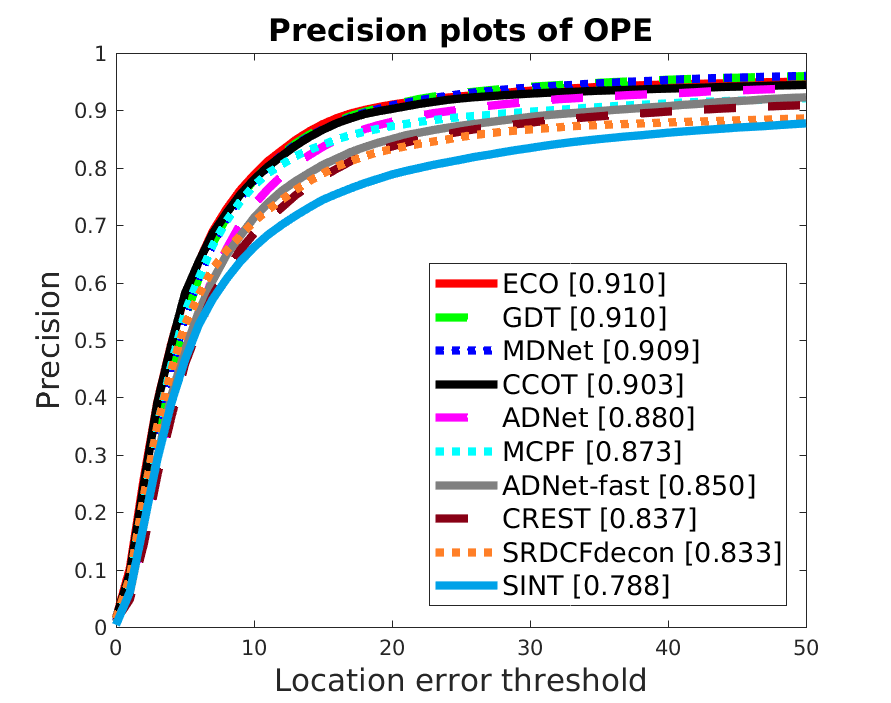}
			
		\end{tabular}

		\caption{Precision and success plots on OTB-2015 using one-pass evaluation.}
		
		\label{fig:2015}
		
	\end{figure}

	\begin{figure}
		\vspace{-0.2in}
		\begin{tabular}{c@{}c@{}c}
			
			\includegraphics[width=0.25\textwidth,height=3.5cm]{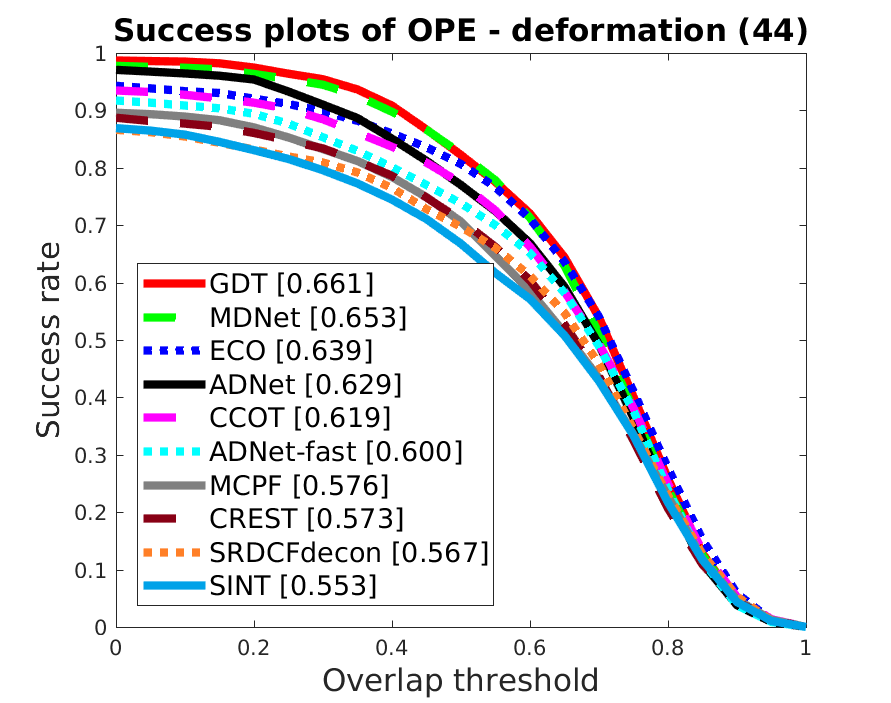}&
			\includegraphics[width=0.25\textwidth,height=3.5cm]{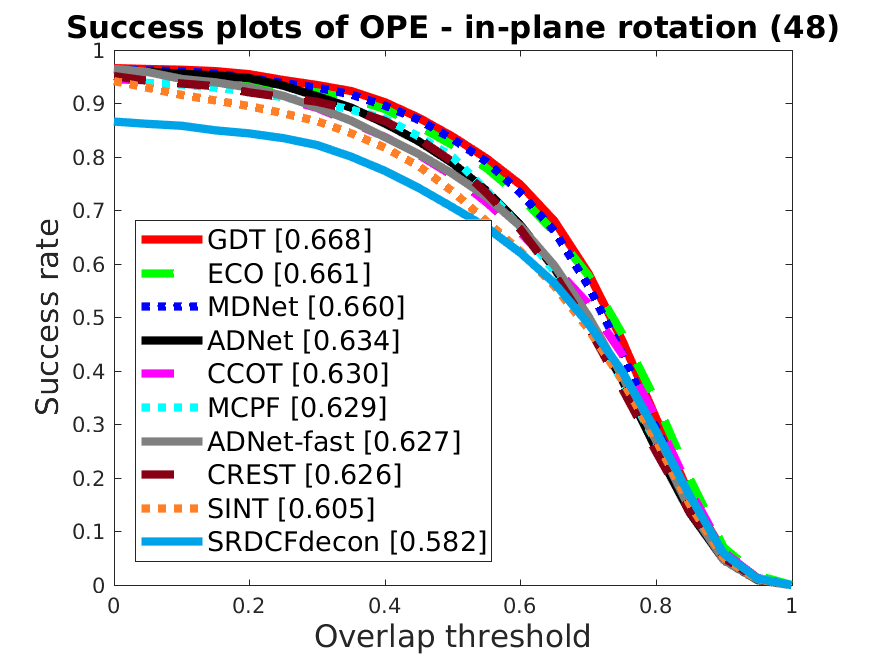}\\
			\includegraphics[width=0.25\textwidth,height=3.5cm]{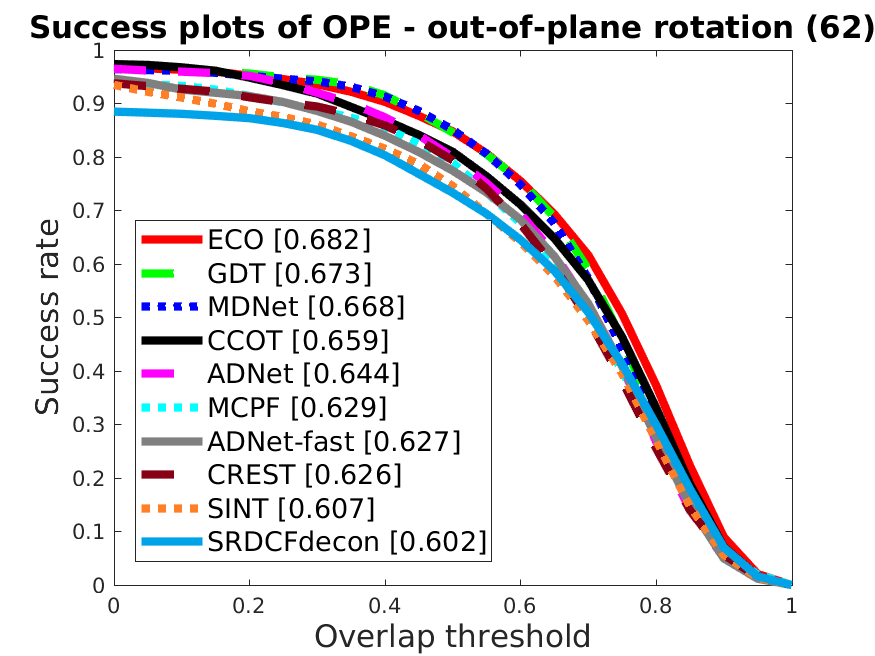}&
			\includegraphics[width=0.25\textwidth,height=3.5cm]{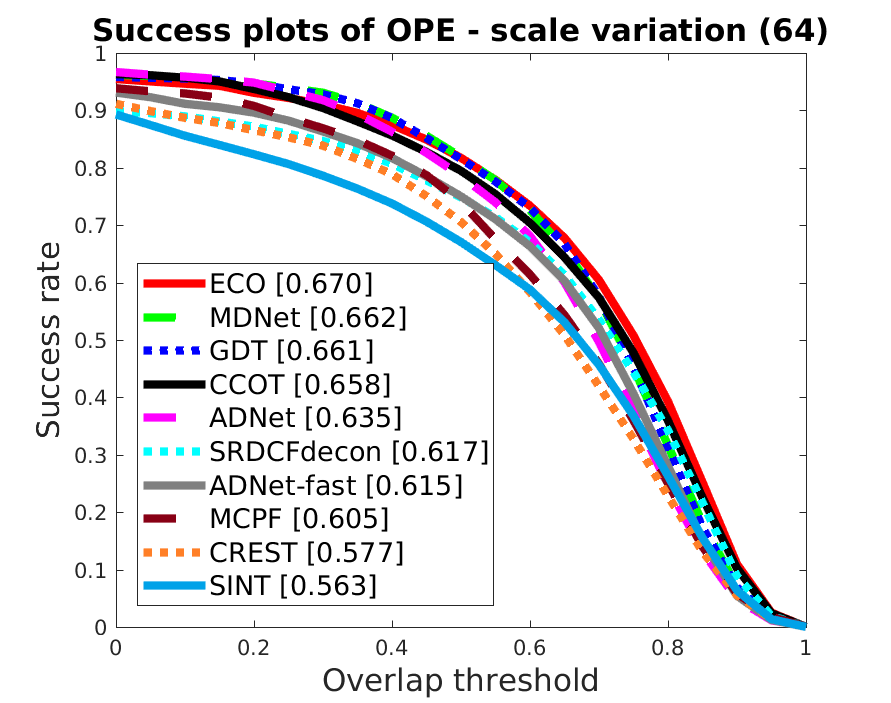}		
		\end{tabular}
		\caption{Overlap success plots over tracking challenges with deformation, in-plane rotation, out-of-plane rotation, and scale variations {on} OTB-2015.}
		
		\label{fig:attr15}
		
	\end{figure}

	\noindent\textbf{OTB-2015 Dataset.} We compare GDT with the state-of-the-art trackers on the OTB-2015 benchmark~\cite{wu2015object}. Fig.~\ref{fig:2015} shows that our proposed tracker GDT performs generally well. Although the ECO tracker achieves the best result in overlap success, GDT performs the same as ECO in distance precision. As far as we know, ECO is the state-of-the-art tracker in several public benchmarks, which improves continuous convolutional operators on a multi-level feature maps.
    Since the OTB-2015 dataset contains more challenging videos with fast motion and low resolution, our tracker fails to match up with ECO {on} success rate. Besides, our method outperforms all of other methods {on} both success rate and precision. In addition, we also illustrate the tracking performance for videos annotated as \textit{deformation, in-plane rotation, out-of-plane rotation}, and \textit{illumination variation} in Fig.~\ref{fig:attr15}. Our model shows advantages in tracking deformable objects and objects with in-plane rotation as well.

    \begin{table}
    	\centering
    	\footnotesize
    	\caption{Comparison with the state-of-the-art trackers on the VOT-2016 dataset. The results are presented in terms of expected average overlap (EAO), accuracy, and robustness. \ryn{Note that the best, second best and third best results are marked in \textit{red/blue/green} colors, respectively. (The same color coding scheme is used in all remaining tables.)}}
    	\setlength{\tabcolsep}{9.5pt}
    	\begin{tabular}{cccccc}
    		\hline
    		& {ECO}   & {CCOT}  & {Staple} & {MDNet} & \textbf{GDT}   \\\hline
    		\textit{EAO} & \textcolor{red}{0.374} & \textcolor{green}{0.331} & 0.295  & 0.257 & \textcolor{blue}{0.353} \\
    		\textit{Accuracy} & \se{0.555} & 0.541 & \textcolor{green}{0.547} & {0.542} & \textcolor{red}{0.585} \\
    		\textit{Robustness} & \textcolor{red}{0.818} & \textcolor{blue}{0.788} & 0.686 & 0.714 & \textcolor{green}{0.774} \\\hline
    	\end{tabular}
    	
    	\label{tab:vot}
    \end{table}

    \begin{figure}[b]
    	\centering
    	\includegraphics[width=0.5\textwidth]{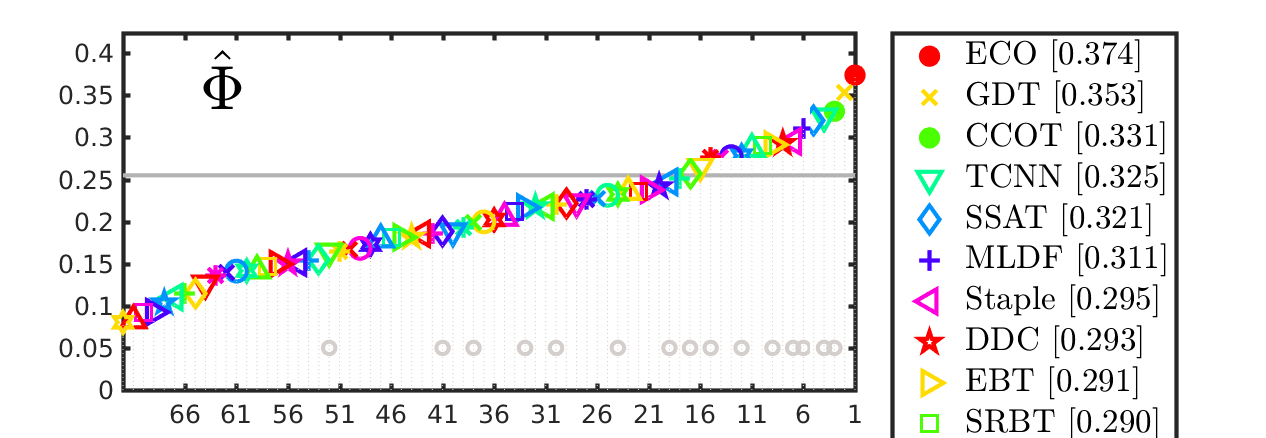}
    	
    	\caption{Comparison with more state-of-the-art trackers on the VOT-2016 dataset. The results are presented specifically in terms of expected average overlap (EAO).}
    	
    	\label{fig:eao}
    	
    \end{figure}



\begin{table}[]
\footnotesize
\centering
\caption{Tracking accuracy (A) and robustness (R) on the VOT-2016 dataset. The attributes tested include camera change (Cam.), illumination change (Illu.), motion change (Mot.), occlusion (Occ.), size change (Size), and not assigned (N/A).}
\setlength{\tabcolsep}{5.5pt}
\label{tab:vot16_1}
\begin{tabular}{ccccccccc}
\hline
&        & Cam.  & Illu. & Mot.  & Occ.  & Size  & N/A   & All   \\\hline

\multirow{5}{*}{\textbf{A}} & ECO    & \fir{0.576} & 0.646 & 0.500 & 0.408 & \se{0.519} & \se{0.595} & \se{0.555} \\

& CCOT   & \thi{0.558} & \thi{0.656} & 0.475 & \thi{0.437} & 0.507 & 0.578 & 0.541 \\

& Staple & 0.555 & \fir{0.715} & \se{0.511} & \thi{0.437} & \thi{0.516} & \thi{0.583} & \thi{0.547} \\

& MDNet  & 0.547 & 0.639 & \thi{0.508} & \se{0.491} & 0.511 & 0.563 & 0.542 \\

& \textbf{GDT}    & \se{0.566} & \se{0.667} & \fir{0.558} & \fir{0.536} & \fir{0.557} & \fir{0.610} & \fir{0.585} \\
\hline\hline

\multirow{5}{*}{\textbf{R}}

& ECO    		& \fir{0.834} & \fir{0.904} & \fir{0.708} & 0.490 & \se{0.817} & \fir{0.932} & \fir{0.818} \\
& CCOT   		& \se{0.748} & \thi{0.818} & \thi{0.666} & \se{0.555} & 0.769 & \thi{0.857} & \se{0.788} \\
& Staple 		& 0.662 & 0.495 & 0.491 & 0.365 & 0.739 & 0.833 & 0.686 \\
& MDNet  		& 0.671 & 0.683 & 0.647 & \fir{0.582} & \thi{0.784} & 0.771 & 0.714 \\
& \textbf{GDT}    & \thi{0.707} & \se{0.857} & \se{0.688} & \thi{0.521} & \fir{0.850} & \se{0.872} & \thi{0.774}\\ \hline

\end{tabular}

\end{table}

    \noindent\textbf{VOT-2016 Dataset.} We compare {GDT} with state-of-the-art trackers on the VOT-2016 benchmark, including ECO~\cite{danelljan2017eco}, Staple~\cite{bertinetto2016staple}, MDNet~\cite{nam2016learning}, and CCOT~\cite{danelljan2014accurate}. Table~\ref{tab:vot} shows that GDT performs comparably to the state-of-the-art trackers on the EAO metric. GDT only performs worse than ECO {on} EAO and ranks at the top in accuracy. Considering the slight loss of robustness induced by the deformable convolution, our GDT's robustness is behind ECO and CCOT.
    In Fig.~\ref{fig:eao}, we analyze EAO by comparing {GDT} with 41 other trackers, including ECO, CCOT, TCNN, Staple, and EBT. GDT has the second highest expected average overlap among them. Since VOT-2016 consists of many short-term challenging tracking scenarios with severe appearance deformation ({e.g.,} \textit{motorcross1}, \textit{gymnastics1}), our approach outperforms most of {the} other methods in these scenarios.

    \ryn{In Tab.~\ref{tab:vot16_1}, we analyze the performance of these trackers on videos with different attributes based on the accuracy and robustness metrics. The annotated attributes include camera change (Cam.), illumination change (Illu.), motion change (Mot.), occlusion (Occ.), size change (Size), and not assigned (N/A). For accuracy, our approach has obvious advantages for most attributes. Although GDT is less robust than CCOT and ECO, we notice that with regard to the size change attribute, our method outperforms the others. }
   	\ryn{Tab.~\ref{tab:vot16_2} shows the results of the unsupervised (non-reset) experiment. Our approach ranks the top on the overall performance and three attributes. Among them, occlusion and size change are closely related to  deformable object tracking, in which our approach gains obvious advantages.}
   	
   	\begin{table}
   		\centering
   		\footnotesize
   		\caption{Comparison with the state-of-the-art trackers on the VOT-2017 dataset. The results are presented in terms of expected average overlap (EAO), accuracy, and robustness.}
   		\label{tab:vot17}
   		\setlength{\tabcolsep}{9.5pt}
   		\begin{tabular}{cccccc}
   			\hline
   			& {ECO}   & {CCOT}  & {Staple} & {SiamFC} & \textbf{GDT}   \\\hline
   			\textit{EAO} & \textcolor{red}{0.282} & \textcolor{blue}{0.269} & 0.175  & 0.187 & \textcolor{green}{0.258} \\
   			\textit{Accuracy} & 0.490 & 0.501 & \textcolor{blue}{0.529} & \textcolor{green}{0.504} & \textcolor{red}{0.558} \\
   			\textit{Robustness} &  \textcolor{red}{0.753} & \textcolor{blue}{0.717} & 0.511 & 0.547 & \textcolor{green}{0.645}\\\hline
   		\end{tabular}
   	\end{table}
   	
   	\begin{table}
   		\footnotesize
   		\centering
   		\caption{Tracking accuracy (A) and robustness (R) on the VOT-2017 dataset. The attributes tested include camera change (Cam.), illumination change (Illu.), motion change (Mot.), occlusion (Occ.), size change (Size), and not assigned (N/A).}
   		\label{tab:vot17_1}
   		\setlength{\tabcolsep}{5.5pt}
   		\begin{tabular}{lcccccccc}
   			\hline
   			&        & Cam.  & Illu. & Mot.  & Occ.  & Size  & N/A   & All   \\			\hline
   			
   			\multirow{5}{*}{\textbf{A}}
   			& ECO    		& 0.513 & \fir{0.519} & 0.481 & 0.339 & 0.446 & 0.497 & 0.490 \\
   			
   			& CCOT   		& \thi{0.516} & 0.478 & \thi{0.488} & \thi{0.391} & 0.451 & 0.513 & 0.501 \\
   			
   			& Staple 		& \se{0.554} & 0.483 & \se{0.508} & \se{0.431} & \se{0.491} & \se{0.534} & \se{0.529} \\
   			
   			& MDNet  		& 0.513 & \se{0.510} & 0.498 & 0.379 & \thi{0.475} & \thi{0.523} & \thi{0.504} \\
   			
   			& \textbf{GDT}    & \fir{0.556} & \thi{0.492} & \fir{0.548} & \fir{0.477} & \fir{0.546} & \fir{0.561} & \fir{0.558} \\			\hline			\hline
   			
   			\multirow{5}{*}{\textbf{R}}
   			& ECO    		& \fir{0.726} & \se{0.390} & \fir{0.621} & \se{0.364} & \fir{0.767} & \fir{0.840} & \fir{0.753} \\
   			
   			& CCOT   		& \se{0.717} & 0.244 & \thi{0.587} & \thi{0.347} & \thi{0.663} & \se{0.789} & \se{0.717} \\
   			
   			& Staple 		& 0.452 & \thi{0.308} & 0.469 & 0.273 & 0.607 & 0.585 & 0.511 \\
   			
   			& MDNet  		& \thi{0.599} & \thi{0.308} & 0.308 & 0.215 & 0.479 & 0.697 & 0.547 \\
   			
   			& \textbf{GDT}    & 0.567 & \fir{0.697} & \se{0.609} & \fir{0.398} & \se{0.745} & \thi{0.720} & \thi{0.645} \\			\hline
   			
   		\end{tabular}
   	\end{table}	
   	\begin{table}[]
   		\centering
   		\footnotesize
   		\caption{The average overlap (AO) for unsupervised experiments on VOT-2016 dataset. Attributes include camera change (Cam.), illumination change (Illu.), motion change (Mot.), occlusion (Occ.), size change (Size), and not assigned (N/A).}
   		\label{tab:vot16_2}
   		\setlength{\tabcolsep}{5pt}
   		\begin{tabular}{cccccccc}
   			\hline
   			& Cam.   & Illu.  & Mot.   & Occ.   & Size   & N/A    & All    \\	\hline
   			
   			ECO    & \thi{0.471} & \fir{0.629} & 0.382 & \thi{0.332} & 0.420 & 0.458 & 0.441 \\
   			
   			CCOT   & \fir{0.484} & \thi{0.594} & \thi{0.397} & 0.328 & \thi{0.461} & \fir{0.510} & \se{0.470} \\
   			
   			Staple & 0.419 & 0.559 & 0.355 & 0.235 & 0.388 & 0.408 & 0.390 \\
   			
   			MDNet  & 0.449 & 0.577 & \se{0.433} & \se{0.341} & \se{0.473} & \se{0.503} & \thi{0.458} \\
   			
   			\textbf{GDT}    & \se{0.475} & \se{0.621} & \fir{0.479}  & \fir{0.391}  & \fir{0.513} & \thi{0.484} & \fir{0.484} \\	\hline
   		\end{tabular}
   	\end{table}
   	
   	\begin{table}
   		\footnotesize
   		\centering
   		\caption{The average overlap (AO) for unsupervised experiments on VOT-2017 dataset. Attributes include camera change (Cam.), illumination change (Illu.), motion change (Mot.), occlusion (Occ.), size change (Size), and not assigned (N/A).}
   		\label{tab:vot17_2}
   		\setlength{\tabcolsep}{5pt}
   		\begin{tabular}{cccccccc}
   			\hline
   			& Cam. &  Illu. & Mot. & Occ. & Size & N/A & All \\\hline
   			
   			{ECO} & \textcolor{blue}{0.4190} & \textcolor{red}{0.420} & \textcolor{blue}{0.362} & \textcolor{green}{0.261} & \textcolor{blue}{0.370} & \textcolor{red}{0.419} & \textcolor{blue}{0.403} \\
   			{CCOT} & \textcolor{green}{0.400} & \textcolor{blue}{0.367} & 0.332 & \textcolor{blue}{0.263} & \textcolor{green}{0.354} & \textcolor{blue}{0.417} & \textcolor{green}{0.392} \\
   			{Staple} & 0.398 & 0.269& \textcolor{green}{0.356}& 0.180 & 0.332 & 0.305 &	0.337\\
   			{SiamFC} & 0.361 & 0.317& 	0.312	& 0.191 &	0.334 &	0.359 &0.343 \\
   			\textbf{GDT} & \textcolor{red}{0.438} & \textcolor{green}{0.341} & \textcolor{red}{0.432} & \textcolor{red}{0.300} & \textcolor{red}{0.482} & \textcolor{green}{0.375} & \textcolor{red}{0.410}\\\hline
   		\end{tabular}
   	\end{table}	

    {\noindent\textbf{VOT-2017 Dataset.} We compare {GDT} with state-of-the-art trackers on the VOT-2017 benchmark, including ECO~\cite{danelljan2017eco}, Staple~\cite{bertinetto2016staple}, SiamFC~\cite{bertinetto2016fully}, and CCOT~\cite{danelljan2014accurate}.
    Overall, our method does not perform better than ECO and CCOT on EAO and robustness, but ranks top on the accuracy, since the deformable convolution increases tracking accuracy but slightly reduces the robustness.}

    \ryn{In Tab.~\ref{tab:vot17_1}, we analyze the performance of these trackers on different videos based on the accuracy and robustness metrics. Similar to the VOT-2016 results, our approach also performs better for most attributes. Due to the gating module, our approach performs well the in illumination change, occlusion, and size change attributes.}

 	\ryn{Tab.~\ref{tab:vot17_2} shows the results of the unsupervised experiment. The annotated attributes are the same as in VOT-2016.
 	Our approach obtains the highest overall AO and outperforms the others in most attributes. As observed, our approach exhibits obvious advantages for the size change attribute, which is most related to deformable object tracking. In particular, GDT has a $30.2\%$ margin over ECO on the Average Overlap metric.}

	\subsection{Qualitative Evaluation}
\begin{figure*}
		\centering
		
		\includegraphics[width=1.0\textwidth]{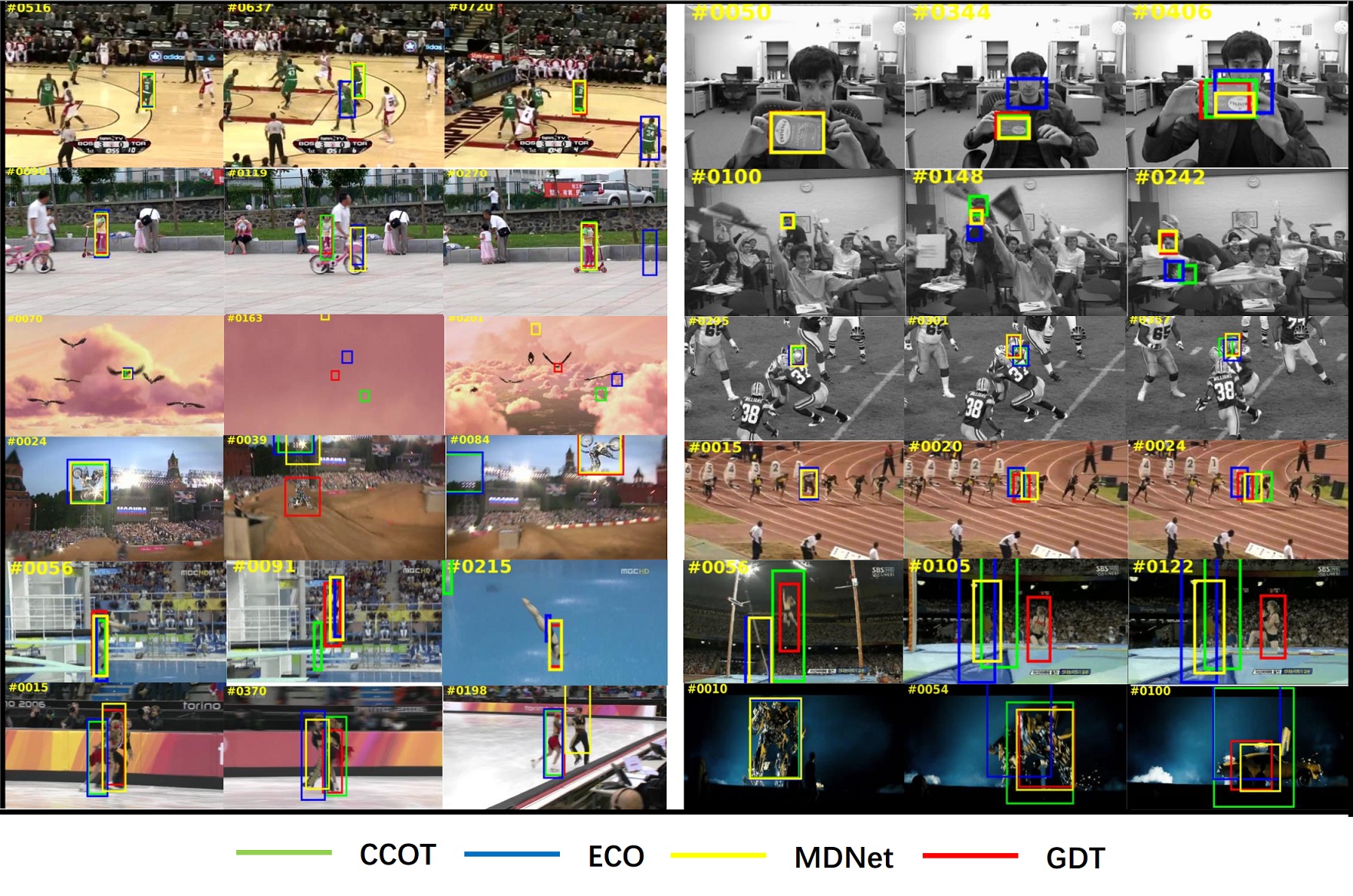}
		
		\caption{Qualitative evaluation of GDT, CCOT [14], MDNet [41] and ECO [9] on 12 challenging
			sequences (from left to right and top to down: \textit{Basketball}, \textit{ClifBar}, \textit{Girl2}, \textit{Freeman1}, \textit{Bird1}, \textit{Football}, \textit{MotorRolling}, \textit{Bolt2}, \textit{Diving},
			\textit{Jump}, \textit{Skater2} and \textit{Trans}) from OTB-2015. Our proposed tracker outperforms the state-of-the-art methods. }
		
		\label{fig:res}
		
	\end{figure*}

	Fig.~\ref{fig:res} visually compare the
	results of the top-performing trackers: CCOT~\cite{danelljan2014accurate}, MDNet~\cite{nam2016learning}, ECO~\cite{danelljan2017eco} and our GDT on 12 challenging sequences.
    In the scenarios of \textit{Diving}, \textit{Jump}, \textit{Skater2}, and \textit{Trans}, which are close-up sequences from sports and movies, the objects exhibit severe pose variations. These results demonstrate the superiority of GDT qualitatively compared to the state-of-the-art trackers.
    For {example}, in \textit{Diving}, CCOT loses track in the early stage and ECO fails to cover the body of the object. MDNet has a slight deviation at frame \#91, while our tracker performs well in the entire sequence. Similarly in \textit{Jump}, MDNet, CCOT, and ECO lose track and they cannot estimate the object size well. Since this sequence is very challenging, our tracker cannot estimate the object scale accurately either, but it still manages to track the person. In the sequence \textit{Skater2} where two skaters frequently interact, the trackers tend to be distracted by the partner of the tracking target, while GDT always clings to the target.
    Both CCOT and ECO extract CNN features and learn correlation filters separately, {and are} not based on an end-to-end architecture. {Hence,} their features are not robust against deformation. MDNet also cannot track an object with severe deformation due to the lack of a mechanism to handle deformation. In the scenario \textit{ClifBar} and \textit{MotorRolling} where the object rotates, ECO and MDNet perform poorly. CCOT also drifts away in \textit{MotorRolling}. \ryn{There are some other examples (\textit{Basketball}, \textit{Girl2}, \textit{Football}, \textit{Freeman1}) with partial occlusions, in which MDNet does not handle well.
    With the aid of the deformable convolution, our tracker is better in adapting to the various object scales and tracking objects with rotations and partial occlusions. }


	\section{Conclusion}
	
	We have proposed a deformable convolution layer to model target appearance variations in the CNN-based tracking-by-detection framework. We aim to capture target appearance variations via deformable convolution and supplement its normal convolution features through the online learned gating module. The gating module controls how the deformable convolutional features and the normal features are fused. Experimental results show that the proposed tracker performs favorably against the state-of-the-art methods.
	
	\begin{figure}
		\vspace{-.1in}
		\begin{tabular}{c@{}c@{}c}
			\includegraphics[width=0.16\textwidth,height=2.9cm]{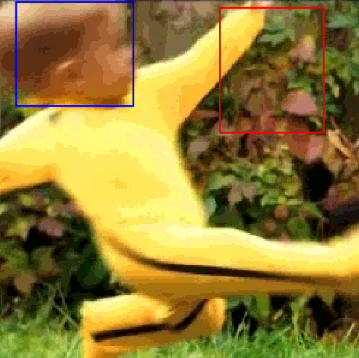} &
			\includegraphics[width=0.16\textwidth,height=2.9cm]{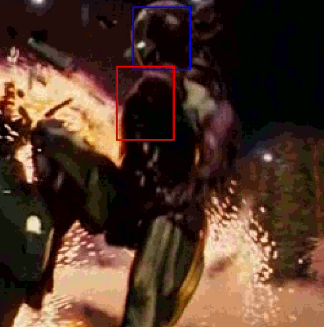}&
			\includegraphics[width=0.16\textwidth,height=2.9cm]{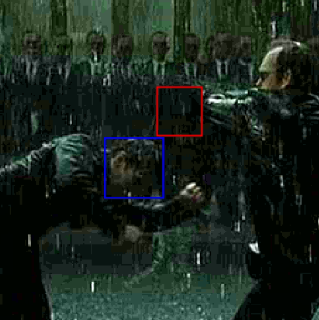}\\
			\footnotesize (a) \textit{DragonBaby} & \footnotesize(b) \textit{Ironman} & \footnotesize(c) \textit{Matrix}\\
		\end{tabular}
		\caption{Examples of failure cases. The ground-truth and the tracking bounding boxes are denoted in blue and red, respectively.}
		
		\label{fig:fail}
		
	\end{figure}

	There are still limitations in our proposed model. Our deformable convolution slightly degrades the robustness of the model, since its deformation estimation may fail in some extreme situations, including long-term occlusion, fast and large deformation, or significant illumination variation, {as shown in Fig.~\ref{fig:fail}}.
	For the future work, we \ryn{would like to improve} the robustness of the deformable convolution by \ryn{enhancing the features extraction stage of our framework. This can be accomplished by collecting more data or adopting a data augmentation technique (e.g., image warping) to independently train the deformable convolution module. In addition, the online learned gating module may not be adequately adaptive to difficult videos. To alleviate this problem, we aim to improve the gating module in the offline training stage. Moreover, in the future, we will consider to generalize our approach to different sources of image data, e.g., RGB-D data and medical images.}
	

	
	%

	%



	\ifCLASSOPTIONcaptionsoff
	\newpage
	\fi

	
	
	%
	\bibliographystyle{IEEEtran}
	\bibliography{egbib}

\vspace{1in}
	

	%
	
	
	
	
	
	

\end{document}